%% file: main.tex
\documentclass[10pt,twocolumn,letterpaper]{article}

\usepackage{cvpr} 

\input{preamble}

\definecolor{cvprblue}{rgb}{0.21,0.49,0.74}
\usepackage[pagebackref,breaklinks,colorlinks,allcolors=cvprblue]{hyperref}

\title{Training-Free Multi-Concept LoRA Composition with Prompt-Aware Weighting}

\author{
    \large Georgios Tsoumplekas$^1$, Stella Bounareli, Vasileios Argyriou$^1$ \\
    $^1$ Department of Networks and Digital Media, Kingston University London, UK
}

\begin{document}
\maketitle

\input{main_sections/00_abstract}    

\input{main_sections/01_introduction}

\input{main_sections/02_related_work}

\input{main_sections/03_method}

\input{main_sections/04_results}

\input{main_sections/05_conclusions}

{
    \small
    \bibliographystyle{ieeenat_fullname}
    \bibliography{main}
}

\clearpage

\appendix
\renewcommand{\thesection}{\Alph{section}}
\setcounter{section}{0}

\input{supplementary_sections/new_metrics_discussion}

\input{supplementary_sections/experimental_results_extras}

\input{supplementary_sections/ablation_study_extras}

\input{supplementary_sections/mllm_evaluation}

\input{supplementary_sections/user_study_extras}

\input{supplementary_sections/limitations_error_cases}

\input{supplementary_sections/future_work}

\input{supplementary_sections/societal_impact}

\input{supplementary_sections/qualitative_comparisons_extras}

\end{document}

%% file: preamble.tex
\newcommand{\TODO}[1]{\textbf{\color{red}[TODO: #1]}}
\renewcommand{\TODO}[1]{}

\usepackage[utf8]{inputenc}
\usepackage[T1]{fontenc}
\usepackage{url}
\usepackage{booktabs}
\usepackage{amsfonts}
\usepackage{nicefrac}
\usepackage{microtype}
\usepackage{graphicx}
\usepackage{graphics}
\usepackage{epsfig}
\usepackage{mathptmx}
\usepackage{times}
\usepackage{amssymb}
\usepackage{amstext}
\usepackage{amsmath}
\usepackage{amsthm}
\usepackage{multirow}
\usepackage{bm}
\usepackage{xcolor}
\usepackage{svg}
\usepackage[most]{tcolorbox}
\usepackage{caption}
\usepackage{natbib}

\lstdefinestyle{promptstyle}{
  basicstyle=\ttfamily\footnotesize,
  columns=fullflexible,
  breaklines=true,
  breakatwhitespace=true,
  keepspaces=true,
  showstringspaces=false,
  frame=none
}

\newtcolorbox{promptbox}[2][]{%
  enhanced,
  colback=blue!3,
  colframe=blue!40,
  boxrule=0.6pt,
  arc=2pt,
  left=6pt,right=6pt,top=6pt,bottom=6pt,
  title=\textbf{#2},
  fonttitle=\small,
  #1
}

%% file: main_sections/00_abstract.tex
\begin{abstract}
Low-Rank Adaptation (LoRA) successfully enables personalization in text-to-image generation by adapting pre-trained diffusion models to specific visual concepts and styles. However, extending such models to multi-concept customization remains challenging. Naively combining multiple LoRA weights or their outputs often leads to interference among concepts, resulting in degraded visual quality and reduced fidelity to the reference images of individual concepts. This paper proposes a simple yet effective approach for multi-concept customization by optimally combining the outputs of multiple LoRA modules. We leverage the relative importance of each concept during generation, as inferred from its corresponding prompt tokens and introduce two methods, W-Switch and W-Composite, that employ a prompt-aware importance weighting strategy in which each LoRA is weighted according to the semantic influence of its trigger words in the target prompt. In addition, we extend existing quantitative evaluation metrics by proposing a new image-based similarity evaluation framework that assesses image fidelity and identity preservation through comparisons between real-world reference images and automatically segmented concept regions from generated images. We evaluate our approach on the ComposLoRA testbed and demonstrate consistent improvements over existing state-of-the-art methods in terms of visual quality, identity preservation and compositionality. Qualitative evaluations, including a Large Language Model (LLM) based assessment and a user study, further validate the effectiveness of the proposed methods and align with the newly introduced quantitative image-based metrics. Our code is available at https://github.com/GeorgeTsoumplekas/Prompt-Aware-Multi-LoRA-Composition.
\end{abstract}

%% file: main_sections/01_introduction.tex
\section{Introduction}
\label{sec:introduction}

Diffusion models (DMs) have emerged as a leading paradigm for both image~\cite{baldridge2024imagen, podell2023sdxl, ramesh2021zero, saharia2022photorealistic} and video~\cite{ho2022video, kong2024hunyuanvideo, singer2022make} generation, with latent diffusion~\cite{rombach2022high} and transformer-based~\cite{peebles2023scalable} architectures demonstrating strong out-of-the-box performance across a wide range of benchmarks. In recent years, the customization of DMs has attracted increasing attention as a means of adapting pre-trained generators to specific visual concepts, styles and downstream tasks~\cite{ruiz2023dreambooth}. In this context, low-rank adaptation (LoRA)~\cite{hu2022lora} has emerged as a dominant approach for customizing text-to-image DMs, offering a computationally efficient solution by optimizing a small set of low-rank matrices within the model layers.

Nonetheless, existing LoRA-based adaptation approaches are typically limited to customizing a single concept. In contrast, real-world images often comprise a mosaic of multiple elements, rendering compositionality~\cite{liu2022compositional} a critical requirement for controllable image generation. Consequently, increasing attention has been directed toward multi-concept customization of text-to-image DMs~\cite{gu2023mix, simsar2025loraclr, zhongmulti}, which enables the simultaneous control of multiple independently learned concepts within a single generation. Such capabilities are critical for applications including virtual try-on systems~\cite{morelli2023ladi}, story-driven image generation~\cite{wang2025characonsist} and realistic modeling of human–object or human–scene interactions~\cite{hoe2024interactdiffusion}.

Extending single-concept customization to a multi-concept setting is non-trivial, as naively merging the weights or outputs of multiple LoRA modules often fails to preserve high performance and fidelity across all concepts. This issue, commonly referred to as interference~\cite{po2024orthogonal, roy2025multlfg}, has been widely observed in multi-concept customization scenarios~\cite{gu2023mix, simsar2025loraclr}. Existing approaches address interference in multi-concept customization either by merging the weights of multiple LoRA adapters into a single adapter~\cite{chen2025iteris, gu2023mix, po2024orthogonal, simsar2025loraclr} or by combining the noise predictions of different LoRAs at each diffusion timestep~\cite{foteinopoulou2025loratorio, roy2025multlfg, zhongmulti, zou2025cached}. The latter strategy is training-free, computationally more efficient and has been shown to be more effective at mitigating interference~\cite{zhongmulti}.

Specifically,~\cite{zhongmulti} is among the first works to adopt this decoding-centric paradigm, introducing LoRA-Switch and LoRA-Composite. In LoRA-Switch, a single LoRA is activated at each diffusion timestep, with all LoRAs scheduled in a periodic manner throughout the sampling process. In contrast, LoRA-Composite computes the final prediction at each timestep by averaging the outputs of all LoRAs. However, while both methods alleviate interference, they represent two extremes of a broader design space, in which the contributions of different LoRA outputs can be weighted unequally or activated over varying numbers of diffusion timesteps. Moreover, both approaches underutilize the influence of the target prompt itself on the generation process.

Additionally, existing quantitative evaluation protocols~\cite{foteinopoulou2025loratorio, roy2025multlfg, zou2025cached} for multi-concept DM customization primarily rely on measuring the semantic similarity between the target prompt and the corresponding generated image. However, improved performance under such metrics does not necessarily correspond to higher image quality, which depends on fidelity at both high-level semantics and low-level visual attributes with respect to real images of the target concepts. This limitation becomes particularly pronounced for images containing human characters, where identity preservation is crucial and necessitates the use of specialized metrics that explicitly measure identity consistency~\cite{deng2019arcface}.

In this paper, we address the underutilization of target prompt semantics during generation by proposing a prompt-aware formulation that adaptively modulates the contributions of multiple LoRA modules according to the semantic importance of their associated prompt tokens. In addition, we address the limitations of existing image-based similarity evaluation protocols by introducing a new similarity evaluation framework that enables assessment of generated images through direct comparison of individual visual concepts with real-world reference images. The main contributions of this paper can be summarized as follows:

\begin{itemize}
    \item We propose \emph{W-Switch} and \emph{W-Composite}, which introduce a simple yet effective mechanism for determining the relative importance of each contributing LoRA during generation based on the semantic influence of their associated trigger words in the target prompt. To the best of our knowledge, this prompt-aware weighting strategy has not been previously explored in the context of decoding-centric multi-concept text-to-image generation.
    \item We extend the evaluation of generated images by proposing a novel similarity evaluation pipeline that assesses image fidelity and identity preservation through comparisons between real-world concept images and cropped concepts from generated images using CLIP~\cite{radford2021learning}, DINO~\cite{oquab2023dinov2}, and ArcFace~\cite{deng2019arcface}.
    \item We achieve state-of-the-art performance on the \emph{ComposLoRA} testbed across existing quantitative benchmarks and our newly introduced image-based metrics. These gains are further supported by improved human preference in visual quality and identity preservation across diverse human characters, as evidenced by both a Large Language Model (LLM) based evaluation and a user study.
\end{itemize}

%% file: main_sections/02_related_work.tex
\section{Related Work}
\label{sec:related_work}

\subsection{Multi-Concept Text-to-Image Composition}

Image compositionality plays a vital role in image generation, particularly in the context of realistic digital content creation. Early approaches to improving compositionality focused on combining the energy functions of different concepts using logical composition operators~\cite{du2020compositional}. Meanwhile, a series of methods for incorporating multiple concepts into text-to-image generation with minimal interference focus on jointly fine-tuning the base DM across all target concepts, enabling it to generate images under multi-concept customization settings~\cite{bill2025jedi, kumari2023multi, liu2023customizable, peng2025tara, voynov2023p+}. While these methods enable multi-concept composition, they lack the computational efficiency required to scale to an increasing number of concepts, as additional fine-tuning is needed for each new concept, making them impractical for modern large-scale text-to-image DMs.

Following a different direction, instead of fine-tuning a single model on all concepts jointly, Kwon et al.~\cite{kwon2024concept} propose combining the outputs of separately fine-tuned models using region masks extracted from the target prompt. Finally, FastComposer~\cite{xiao2025fastcomposer} proposes augmenting the generic text conditioning in DMs with subject embeddings extracted by an image encoder, enabling multi-concept generation at inference time. However, these methods still require extensive training, making them computationally expensive and limiting their scalability as the number of custom concepts increases.

\subsection{Merging Multiple LoRA Models}

In the context of skill compositionality, merging multiple LoRA modules has enabled the composition of diverse skills in large base models, including LLMs and DMs while minimizing the cost of additional fine-tuning. More broadly, a growing body of work has explored skill composition for LLMs and foundation models in downstream tasks~\cite{huanglorahub, prabhakar2025lora, zheng2025decouple}.

A well-studied approach in this context is the direct merging of LoRA module weights, which has gained popularity for content and style adaptation. This line of work includes both training-free methods~\cite{ouyang2025k, zhang2025subject} and training-based approaches~\cite{chen2025consislora, frenkel2024implicit, li2025autolora, shah2024ziplora, shenaj2025lora, yang2025qr}. However, these approaches are typically limited to combining only two LoRA modules, resulting in a narrow formulation compared to the broader problem of multi-concept customization.

Subsequently, a range of approaches has been proposed to extend direct weight merging to the multi-concept composition setting~\cite{chen2025iteris, gu2023mix, po2024orthogonal, simsar2025loraclr, zhang2025rethinking}. More recent work further generalizes this direction by reusing principal subspaces of existing LoRA weights to more efficiently learn combined LoRA modules~\cite{kaushik2025eigenlorax}. A key advantage of these approaches is that at each denoising timestep only the output of the merged LoRA is required. However, they tend to exhibit high interference among concepts~\cite{zhongmulti}.

On the other hand, rather than directly merging LoRA modules in weight space, several training-based approaches explicitly focus on learning how to update the DM’s latent variables at each denoising timestep while multiple LoRAs are active~\cite{meral2025contrastive, yang2025lora}. Interference can also be mitigated by applying LoRAs to different spatial regions of the image~\cite{dalva2025lorashop} or by binding and activating each LoRA through distinct subject tokens in the target prompt~\cite{zheng2025freelora}.

Most closely related to our work are decoding-centric, training-free approaches that merge the noise prediction outputs of multiple LoRA modules at each denoising timestep~\cite{zhongmulti}. These methods primarily focus on inferring appropriate weights to determine the contribution of each LoRA output at each timestep, leveraging signals from the spatial domain~\cite{foteinopoulou2025loratorio}, the frequency domain~\cite{roy2025multlfg}, or temporal changes in the generated image induced by each LoRA~\cite{zou2025cached}. Our method aims to substantially simplify the determination of LoRA contribution weights by leveraging the semantic influence of each LoRA’s associated trigger words in the target prompt. While prompt-based strategies have been explored in the context of multi-concept customization~\cite{liu2023customizable, xiao2025fastcomposer, zheng2025freelora}, their application within the decoding-centric training-free framework remains underexplored.

%% file: main_sections/03_method.tex
\section{Proposed Method}
\label{sec:proposed_method}

\subsection{Preliminary}

\paragraph*{Latent Text-to-Image Diffusion Models}

Text-to-image DMs are built upon denoising diffusion probabilistic models (DDPMs)~\cite{ho2020denoising, sohl2015deep, song2020denoising}, which synthesize data by learning to invert a gradual noising process. In this work, we adopt Stable Diffusion (SD)~\cite{rombach2022high} as the base text-to-image DM for all experiments. SD is a latent DM that performs the denoising process in a learned latent space, enabling computationally efficient high-resolution image generation. Moreover, it incorporates textual conditioning by encoding a text prompt $p$ into a semantic embedding $c$ which guides the image generation process.

Given an input image $x_0$, an encoder $\mathcal{E}$ maps it to a latent representation $z_0=\mathcal{E}(x_0)$. The forward diffusion process subsequently corrupts $z_0$ by progressively adding Gaussian noise according to a predefined noise schedule. Specifically, at timestep $t$, the noisy latent $z_t$ is obtained as $z_t = \sqrt{\alpha_t}z_0 + \sqrt{1-\alpha_t} \epsilon$ where $\epsilon \sim \mathcal{N}(0,I)$ and $\{\alpha_t\}_{t=1}^T$ is a monotonically decreasing sequence controlling the noise magnitude.

The denoising network $\epsilon_\theta$, parameterized by $\theta$, is trained to predict the injected noise conditioned on the noisy latent $z_t$, the diffusion timestep $t$ and the text embedding $c$. The training objective minimizes the expected mean-squared error between the predicted noise and the ground-truth noise.

To further enhance the influence of textual conditioning during sampling, SD adopts classifier-free guidance (CFG)~\cite{ho2022classifier}. During training, the model is optimized using both conditional and unconditional objectives by randomly dropping the conditioning signal. At inference time, guidance is applied by combining the conditional and unconditional noise predictions using a guidance scale that controls the strength of conditioning.

\paragraph*{Low-Rank Adaptation}
LoRA~\cite{hu2022lora} is a parameter-efficient fine-tuning technique for adapting large pre-trained models to downstream tasks by updating only a small number of additional parameters. LoRA is motivated by the observation that, during fine-tuning, weight update matrices exhibit a low intrinsic rank~\cite{aghajanyan2021intrinsic}.

Formally, given a pre-trained weight matrix $W_0 \in \mathbb{R}^{m \times n}$ in a neural network, LoRA freezes the original weights and parameterizes the weight update $\Delta W$ using a low-rank decomposition, rather than directly fine-tuning $W_0$. The adapted weight matrix is defined as:

\begin{equation}
    W = W_0 + \Delta W = W_0 + BA,
\end{equation}

where $B \in \mathbb{R}^{m \times r}$ and $A \in \mathbb{R}^{r \times n}$ are trainable matrices, and $r \ll \min(m,n)$ denotes the chosen rank. During training, only the parameters in $A$ and $B$ are optimized, while $W_0$ remains fixed. Due to its efficiency and flexibility, LoRA has been widely adopted for fine-tuning large-scale DMs.

\paragraph*{Decoding-Centric LoRA Merging}

While a single LoRA module typically specializes in modeling a single concept, composing multiple LoRAs for multi-concept customization remains challenging due to semantic interference and instability that arise from naively merging their weights~\cite{gu2023mix}. Among the earliest training-free approaches that explicitly address multi-LoRA composition at inference time are LoRA-Switch and LoRA-Composite~\cite{zhongmulti}, both of which adopt a decoding-centric strategy.

LoRA-Switch activates only one LoRA at each denoising timestep. Given a set of $N$ LoRAs, the denoising process is partitioned into segments of length $\tau$ and the active LoRA is periodically switched across timesteps. Formally, at denoising step $t$, the index of the active LoRA and the corresponding effective weight matrix are defined as:

\begin{equation} \label{lora_switch_eq}
    \begin{aligned}
        i(t) &= \left\lfloor \frac{(t - 1) \bmod (N\tau)}{\tau} \right\rfloor + 1, \\
        \hat{W}_t &= W + \Delta W{i(t)},
    \end{aligned}
\end{equation}

where $W$ denotes a base model weight matrix and $\Delta W_i = B_i A_i$ represents the low-rank update associated with the $i$-th LoRA. By activating only a single LoRA module at each timestep, LoRA-Switch enforces temporal separation between concepts, allowing each LoRA to influence the generation process in isolation~\cite{zhongmulti}.

In contrast, LoRA-Composite adopts the opposite design choice by simultaneously incorporating all LoRA modules at every denoising step, operating directly at the level of noise prediction. Let $\epsilon_{\theta_i}$ denote the noise predictor of the DM augmented with the $i$-th LoRA. At timestep $t$, LoRA-Composite computes both unconditional and conditional predictions for each LoRA and aggregates them via uniform averaging:

\begin{equation} \label{lora_composite_eq}
    \hat{\epsilon}(z_t, t, c) = \frac{1}{N} \sum_{i=1}^{N} \left[ \epsilon_{\theta_i}(z_t, t) + s \bigl( \epsilon_{\theta_i}(z_t, t, c) - \epsilon_{\theta_i}(z_t, t) \bigr) \right],
\end{equation}

where $s$ denotes the classifier-free guidance scale and all LoRA modules are typically assigned equal weights. This formulation ensures that each LoRA contributes consistently throughout the entire denoising trajectory, promoting balanced semantic integration and improved visual coherence.

LoRA-Switch and LoRA-Composite represent two extremes of decoding-centric multi-LoRA composition, imposing either strict temporal exclusivity or uniform aggregation across all timesteps. Although effective, these fixed formulations limit flexibility by relying on uniform switching mechanisms or by assuming equal, time-invariant contributions from all LoRA modules. In this work, we extend these approaches by introducing a prompt-aware weighting mechanism, in which each LoRA module contributes to the generation process in proportion to the relevance of the concept it represents for a given target prompt.

\subsection{Prompt-based Importance Weighting Mechanism} \label{weighting_mechanisms_section}

\begin{figure}[h]
    \centering
    \includegraphics[width=0.49\textwidth]{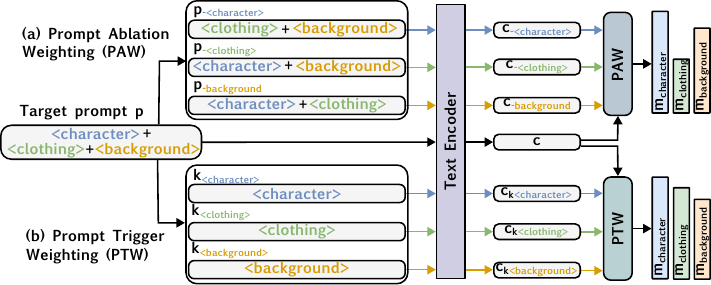}
    \caption{\textbf{Prompt-based relative importance estimation via text-encoder similarity.} Given a target prompt $p$ and per-LoRA trigger-word sets $\mathcal{K}_i$, we compute a relative importance score for each LoRA. \textbf{(a) PAW} removes LoRA-$i$ trigger words $\mathcal{K}_i$ from the prompt to form $p_{-i}$ and scores importance by semantic change $m_i^{\mathrm{PAW}}=1-\cos(c,c_{-i})$. \textbf{(b) PTW} encodes $\mathcal{K}_i$ and scores alignment with the prompt $m_i^{\mathrm{PTW}}=\cos(c,c_{k_i})$.}
    \label{fig:prompt_ablation_methods}
\end{figure}

Our proposed method integrates multiple LoRA modules during diffusion sampling, with each module contributing proportionally to its relevance to the target prompt. To quantify the contribution of each LoRA, we compute similarity-based weights that reflect the semantic influence of the LoRA’s associated trigger words with respect to the target prompt. Specifically, we introduce two novel relative importance weighting mechanisms, denoted as \emph{Prompt Ablation Weighting (PAW)} and \emph{Prompt Trigger Weighting (PTW)}, which differ in how the trigger words of each LoRA are compared against the original target prompt to estimate the LoRA’s relative influence on the generation process. Fig.~\ref{fig:prompt_ablation_methods} illustrates both prompt-based importance weighting mechanisms.

The first weighting strategy, \emph{PAW}, is based on the extent to which a LoRA’s trigger words influence the target prompt, as measured by the semantic change induced when these words are removed. Let $\mathcal{K}_i$ denote the set of trigger words (or keywords) associated with LoRA $i$. Removing these terms from the original prompt $p$ yields a modified prompt $p_{-i} = p \setminus \mathcal{K}_i$. The relative importance score of LoRA $i$ is defined as:

\begin{equation} \label{ablated_eq}
    m_i^{\text{PAW}} = 1-\cos(c, c_{-i}),
\end{equation}

where $c$ and $c_{-i}$ are the text-encoder embeddings of the original prompt $p$ and the modified prompt $p_{-i}$, respectively. As shown in Fig.~\ref{fig:prompt_ablation_methods}(a), larger changes between $c$ and $c_{-i}$ yield higher relative importance.

In contrast, the weighting strategy \emph{PTW} estimates the relative importance of LoRA $i$ by directly measuring the semantic similarity between the original target prompt $p$ and the trigger words associated with that LoRA. Let $c_{k_i}$ denote the text-encoder embedding of the trigger word set $\mathcal{K}_i$ for LoRA $i$. The relative importance score is then defined as:

\begin{equation} \label{triggered_eq}
    m_i^{\text{PTW}} = \cos(c, c_{k_i}).  
\end{equation}

Fig.~\ref{fig:prompt_ablation_methods}(b) shows how \emph{PTW} scores each LoRA via the similarity between the prompt embedding and the embedding of its trigger words.

Both weighting mechanisms rely on text-encoder embeddings and their semantic similarity. A potential concern, particularly for the \emph{PAW} strategy, is that LoRAs with more trigger words may induce a larger discrepancy between $c$ and $c_{-i}$, leading to higher importance scores~\cite{dumpala2024sugarcrepe++}. However, this behavior is desirable as it aligns with human interpretation of textual descriptions. Specifically, concepts described with greater detail are typically more prominent in human understanding. Accordingly, a LoRA characterized by a richer set of trigger words corresponds to a more thoroughly specified concept and, as a result, should exert greater influence during the generation process.

\begin{figure*}[t]
    \centering
    \includegraphics[width=0.8\textwidth]{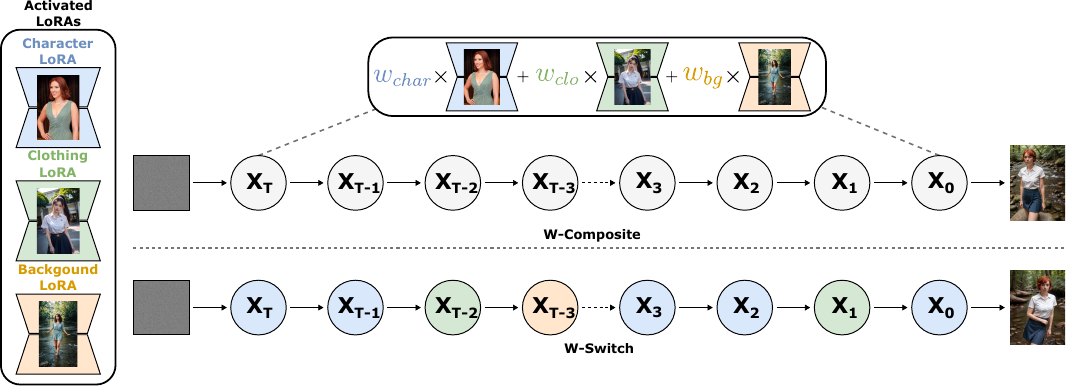}
    \caption{\textbf{Overview of the two prompt-aware multi-LoRA composition methods.} \textbf{W-Composite} aggregates the LoRA-augmented noise predictions at every timestep using fixed, prompt-derived weights $w_i$. \textbf{W-Switch} activates exactly one LoRA per timestep, following a cyclic schedule with within-cycle block lengths proportional to $w_i$.}
    \label{fig:main_methods}
\end{figure*}

\subsection{Weighted Multi-LoRA Composition}

Next, we extend LoRA-Switch and LoRA-Composite~\cite{zhongmulti} to \emph{W-Switch} and \emph{W-Composite}, respectively, by natively integrating the relative importance weights into their underlying mechanisms. Fig.~\ref{fig:main_methods} summarizes the two decoding-centric variants proposed in this work.  Motivated by the intuition that more important LoRAs should exert a stronger influence on the noise prediction at each denoising step, we extend the original weighting mechanism of LoRA-Composite in (\ref{lora_composite_eq}) to \emph{W-Composite} by incorporating normalized relative importance weights obtained via \emph{PAW} or \emph{PTW}. Consequently, this formulation yields a weighted average of the LoRA outputs at each timestep:

\begin{equation} \label{w_composite_eq}
    \hat{\epsilon}(z_t, t, c) = \sum_{i=1}^{N} \left(w_i \times \left[ \epsilon_{\theta_i}(z_t, t) + s \bigl( \epsilon_{\theta_i}(z_t, t, c) - \epsilon_{\theta_i}(z_t, t) \bigr) \right] \right),
\end{equation}

where $N$ is the number of LoRAs used for the given target prompt, $w_i$ denotes the normalized relative importance weight of the $i$-th LoRA, defined as $w_i = m_i / \sum_{j=1}^{N} m_j$, such that $\sum_{i=1}^{N} w_i = 1$, and $m_i \in \{ m_i^{\text{PAW}}, m_i^{\text{PTW}} \}$. As illustrated in Fig.~\ref{fig:main_methods} (top), all LoRAs contribute at every timestep, with constant weights $w_i$.

Similarly, we extend LoRA-Switch to a weighted formulation that enables finer control over the influence of each LoRA throughout the diffusion process. We introduce \emph{W-Switch}, in which only a single LoRA is active at each denoising timestep, while the proportion of timesteps allocated to each LoRA is governed by its associated normalized relative importance weight $w_i$. While in LoRA-Switch all $N$ participating LoRAs are activated for $\tau$ timesteps within each periodic cycle of length $L=N\tau$, as shown in (\ref{lora_switch_eq}), \emph{W-Switch} retains the same cycle length $L$ but allocates within-cycle block lengths proportionally to the normalized importance weights $w_i$. Specifically, we define $q_i = Lw_i$ and construct integer block lengths $\{b_i\}_{i=1}^N$ satisfying $\sum_{i=1}^N b_i = L$ as:

\begin{equation}
    b_i = \lfloor q_i \rfloor + \mathbb{I}[i \in \mathcal{R}], 
    \qquad
    |\mathcal{R}| = L - \sum_{j=1}^{N} \lfloor q_j \rfloor,
\end{equation}

where $\mathcal{R}$ denotes the set of indices corresponding to the  $|\mathcal{R}|$ largest fractional components $q_i - \lfloor q_i \rfloor$. Consequently, we define the cumulative block endpoints as:

\begin{equation}
    l_0 = 0, \quad
    l_i = \sum_{j=1}^{i} b_j \quad (i = 1, \ldots, N),
\end{equation}

such that $l_N=L$. Then, for each denoising step $t \in \{1,\ldots,T\}$ the active LoRA index is determined by the following closed-form expression:

\begin{equation} \label{w_switch_eq}
    \begin{aligned}
        i(t) &= 1 + \sum_{k=1}^{N-1} \mathbb{I}\left(((t - 1) \bmod L) \ge l_k\right), \\
        \hat{W}_t &= W + \Delta W{i(t)}.
    \end{aligned}
\end{equation}

Fig.~\ref{fig:main_methods} (bottom) visualizes the resulting hard switching where the active LoRA changes over time according to the allocated within-cycle blocks. This formulation implements a cyclic schedule that applies $b_1$ consecutive steps of the first LoRA, followed by $b_2$ steps of the second LoRA and so on, up to $b_N$ steps of the $N$-th LoRA, repeating until all $T$ denoising steps are completed. As a result, the $i$-th LoRA is allocated an approximate proportion of $b_i / L$ steps, which converges to the target weight $w_i$ while preserving the strict temporal separation between LoRAs enforced by LoRA-Switch.

While this formulation improves upon LoRA-Switch, empirical results indicate a potential degradation in identity preservation. Since DMs follow a coarse-to-fine generation process~\cite{choi2022perception, park2023understanding} faithful identity preservation critically depends on the final denoising stages. Accordingly, we modify \emph{W-Switch} to prioritize any human-identity LoRA during the last $L_{tail}$ diffusion steps (we set $L_{tail} = 5$) to ensure stronger alignment of facial details between generated and real images.

%% file: main_sections/04_results.tex
\section{Experimental Results}
\label{sec:experimental_results}

\subsection{Experimental Setup}
We follow the experimental protocol of~\cite{zhongmulti, zou2025cached}, adopting SD v1.5~\cite{rombach2022high} as the backbone model and using the Realistic Vision V5.1 checkpoint to facilitate high-fidelity image generation. Unless otherwise specified, all experiments are conducted with $T=100$ denoising steps, a classifier-free guidance scale of $s=7$, and an image resolution of $1024 \times 768$. We employ DPM-Solver++~\cite{lu2022dpm} as the sampling algorithm, and uniformly scale all LoRA modules with a fixed weight of 0.8. Our evaluation focuses on the realistic subset of the \emph{ComposLoRA} benchmark~\cite{zhongmulti}, which comprises a total of 11 LoRA modules, including 3 character LoRAs, 2 background LoRAs, 2 clothing LoRAs, 2 object LoRAs and 2 style LoRAs.

In the following experiments, we evaluate \emph{W-Switch} and \emph{W-Composite}. Based on empirical performance, we adopt \emph{PAW} as the relative importance weighting mechanism for \emph{W-Switch} and \emph{PTW} for \emph{W-Composite}. We compare our proposed methods against LoRA-Switch and LoRA-Composite~\cite{zhongmulti}, as well as CMLoRA~\cite{zou2025cached} which determines per-LoRA contribution weights via a dynamic caching strategy coupled with a dominant weighting scheme.

Following the model settings of~\cite{zhongmulti}, we set the base segment length of each LoRA in \emph{W-Switch} to $\tau = 5$. Since our approach is training-free, all experiments are conducted on a single NVIDIA RTX A6000 GPU and results are reported as the average over three independent runs.

\begin{table*}[t]
\centering
\caption{Comparison of the proposed methods with state-of-the-art baselines on the \emph{ComposLoRA} testbed. Best results are shown in \textbf{bold} and second-best are \underline{underlined}.}
\label{tab:main_results_metrics_max}
\setlength{\tabcolsep}{3pt}
\renewcommand{\arraystretch}{1.15}
\resizebox{\textwidth}{!}{%
\begin{tabular}{lcccccccccccccccccccc}
\toprule
\multirow{2}{*}{Method}
 & \multicolumn{5}{c}{$I_{\mathrm{CLIP}}$}
 & \multicolumn{5}{c}{$I_{\mathrm{DINO}}$}
 & \multicolumn{5}{c}{$I_{\mathrm{ArcFace}}$}
 & \multicolumn{5}{c}{$T_{\mathrm{CLIP}}$} \\
\cmidrule(lr){2-6}
\cmidrule(lr){7-11}
\cmidrule(lr){12-16}
\cmidrule(lr){17-21}
 & N=2 & N=3 & N=4 & N=5 & Avg.
 & N=2 & N=3 & N=4 & N=5 & Avg.
 & N=2 & N=3 & N=4 & N=5 & Avg.
 & N=2 & N=3 & N=4 & N=5 & Avg. \\
\midrule
Switch~\cite{zhongmulti}
 & 75.53 & \underline{74.70} & \underline{73.53} & \underline{72.33} & \underline{74.02} 
 & \underline{53.80} & \textbf{51.88} & \textbf{49.66} & \underline{47.49} & \underline{50.71} 
 & 51.80 & 52.02 & 50.22 & 50.90 & 51.24 
 & \textbf{34.91} & \underline{36.44} & \underline{37.11} & \underline{37.46} & \underline{36.48} \\
Composite~\cite{zhongmulti}
 & 74.35 & 74.35 & 72.89 & 70.18 & 72.94 
 & 50.59 & 50.80 & \underline{49.24} & 45.42 & 49.01
 & \textbf{55.79} & 51.46 & 51.80 & 50.74 & 52.45
 & 32.41 & 35.73 & 35.79 & 35.76 & 34.92 \\
CMLoRA~\cite{zou2025cached}
 & \underline{76.21} & 74.29 & 71.27 & 69.16 & 72.73 
 & 51.39 & 48.12 & 44.31 & 41.05 & 46.22 
 & 52.32 & 50.46 & 48.47 & 48.85 & 50.03 
 & 33.79 & 34.46 & 33.88 & 33.76 & 33.97 \\
 \midrule
W-Composite
 & 76.20 & 74.34 & 72.55 & 70.30	& 73.35	
 & 53.70 & 51.16	& 48.66 & 44.98	& 49.63	
 & \underline{54.11}	& \underline{52.42}	& \underline{52.40}	& \underline{52.05} & \underline{52.75}
 & 34.70	& 35.59	& 35.69	& 35.71	& 35.42 \\
W-Switch
 & \textbf{76.23} & \textbf{75.80} & \textbf{74.72} & \textbf{73.80} & \textbf{75.14}
 & \textbf{54.20} & \underline{51.57} & 49.14 & \textbf{48.03} & \textbf{50.74}
 & 53.57 & \textbf{53.34} & \textbf{52.63} & \textbf{52.68} & \textbf{53.06}
 & \underline{34.78} & \textbf{36.70} & \textbf{37.12} & \textbf{37.55} & \textbf{36.54} \\
\bottomrule
\end{tabular}%
}
\end{table*}

\subsection{Evaluation Metrics} \label{evaluation_metrics_main}

Prior studies on the \emph{ComposLoRA} testbed~\cite{foteinopoulou2025loratorio, zou2025cached} have primarily relied on CLIPScore~\cite{hessel2021clipscore}, which measures the cosine similarity between CLIP~\cite{radford2021learning} embeddings of the target prompt and the generated image. However, recent work~\cite{levi2024double} demonstrates that text and image embeddings occupy distinct manifolds within CLIP’s embedding space, making cross-modal similarity comparisons across models less reliable. Consequently, a more robust evaluation strategy is to compare embeddings of generated images directly against real-world images, as both reside on the same manifold.

Following prior work on LoRA weight merging~\cite{gu2023mix, po2024orthogonal, simsar2025loraclr}, we evaluate generated images by comparing their embeddings against real images using CLIP and DINOv2~\cite{oquab2023dinov2}, yielding the $I_{\text{CLIP}}$ and $I_{\text{DINO}}$ metrics. While these image–image similarity measures mitigate several limitations of CLIPScore (denoted as $T_{\text{CLIP}}$ for consistency), they remain insufficiently aligned with the requirements of our experimental setting. Specifically, we identify two key limitations of these image-based metrics. First, generated images often contain multiple concepts, whereas reference images typically depict a single concept. Directly comparing their global embeddings is therefore suboptimal, as the presence of additional concepts introduces noise into the similarity measurements. Second, per-image scores are typically computed by averaging cosine similarities between a generated image and multiple reference images. This averaging implicitly favors images that lie near the centroid of the reference embedding set, which may correspond to a conceptual mean that does not resemble any realistic instance. In contrast, a generated image that lies close to a specific reference embedding may be more semantically faithful than one that is merely closer to the centroid.

To address these challenges, we introduce a cropping and max-pooling strategy. Specifically, we employ SAM3~\cite{carion2025sam} to localize and crop each concept present in a generated image, using the trigger words associated with the corresponding LoRAs as prompts. For human-related concepts, we apply the FAN face detector~\cite{bulat2017far} to extract individual facial regions. Given a generated image composed of $N$ contributing LoRAs, we extract $N$ concept-specific image crops, each corresponding to a distinct concept. For each concept, we compute the maximum cosine similarity between the embedding of the cropped region and the embeddings of its associated reference images. The final per-image score is then obtained by averaging these maximal similarities across all concepts present in the image:

\begin{equation}
    I_{E}(x) =
    \frac{1}{N}
    \sum_{i=1}^{N}
    \max_{1 \le k \le K_i}
    \cos\!\Big(
    \phi_E\!\big(\mathrm{crop}_i(x)\big),
    \phi_E\!\big(r_{i,k}\big)
    \Big),
\end{equation}

where $x$ denotes a generated image containing $N$ contributing LoRAs, $\mathrm{crop}_i(x)$ corresponds to the image region associated with LoRA $i$ obtained via SAM3 or FAN in the case of human identities. For each concept $i$, $\{r_{i,k}\}_{k=1}^{K_i}$ denotes the set of real reference images used for evaluation and the function $\phi_E(\cdot)$ maps an image to its embedding under encoder $E$ (CLIP or DINO).

Finally, following prior work~\cite{po2024orthogonal, simsar2025loraclr}, we evaluate identity alignment using ArcFace~\cite{deng2019arcface}, yielding the $I_{\text{ArcFace}}$ metric. This metric is computed on the same cropped character images and follows the same formulation as $I_{\text{CLIP}}$ and $I_{\text{DINO}}$. A more detailed discussion of the limitations of existing text-based and image-based alignment metrics, along with an in-depth description of the proposed evaluation pipeline, is provided in Appendix~\ref{sec:metrics_discussion}.

To assess the compositional and aesthetic quality of our approach, we follow~\cite{zou2025cached} and employ MiniCPM-V~\cite{yao2024minicpm} to evaluate compositional image generation along four dimensions: element integration, spatial consistency, semantic accuracy and aesthetic quality, using prompt-guided scores ranging from 0 to 10. In each evaluation round, images generated by different methods are compared under identical prompts and random seeds and the final scores are obtained by averaging across seeds. Additional details of the evaluation protocol are provided in Appendix~\ref{sec:minicpm_evaluation}.

\begin{figure}[t]
  \centering
  \includegraphics[width=\columnwidth]{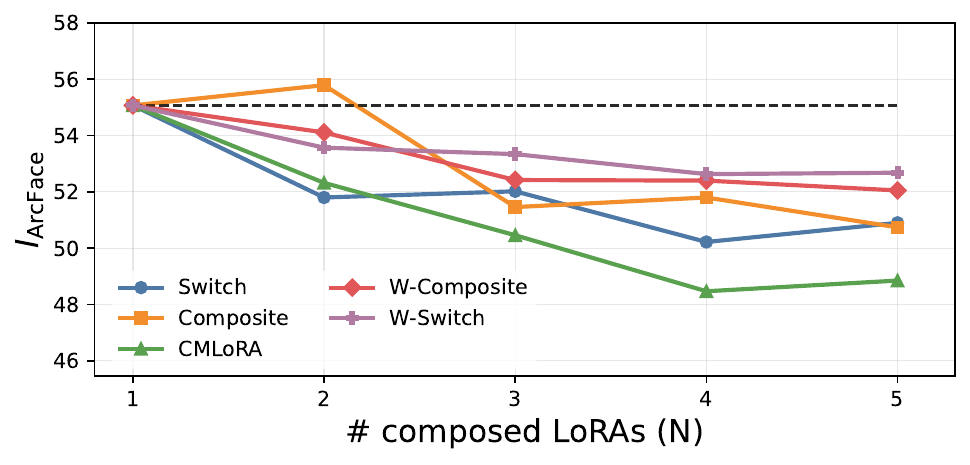}
  \caption{$\mathbf{I_{ArcFace}}$ \textbf{vs. the number of composed LoRAs $\mathbf{N}$.} The dashed line denotes the $N=1$ upper bound. Identity alignment degrades only slightly as more LoRAs are composed.}
  \label{fig:arcface_upper_bound}
\end{figure}

\subsection{Quantitative Results} \label{sec:quantitative_results}

We first report quantitative results on the \emph{ComposLoRA} testbed, following the evaluation protocol of~\cite{zou2025cached}. Table~\ref{tab:main_results_metrics_max} summarizes performance across the proposed image-based metrics $I_{CLIP}$, $I_{DINO}$, and $I_{ArcFace}$ (Section~\ref{evaluation_metrics_main}), together with the text–image alignment metric $T_{CLIP}$ used in prior work~\cite{foteinopoulou2025loratorio, zou2025cached}. Results are reported for varying levels of compositional complexity, with the number of composed LoRAs $N$ ranging from two ($N=2$) to five ($N=5$). Notably, \emph{W-Switch} achieves the best performance across all four metrics. Switch serves as a strong baseline across $I_{DINO}$, $I_{CLIP}$, and $T_{CLIP}$, supporting prior findings~\cite{zhongmulti}. Nonetheless, both \emph{W-Switch} and \emph{W-Composite} consistently outperform their respective vanilla counterparts on average across all evaluated metrics, showing that prompt-aware weighting can improve the generation process.

We further observe that, for the image-based alignment metrics, increasing the number of composed concepts leads to a rapid performance degradation in prior methods such as CMLoRA showing their limited robustness to concept interference. In contrast, both \emph{W-Switch} and \emph{W-Composite} exhibit a substantially slower rate of decline as $N$ increases, indicating improved robustness when generating images with a larger number of customized concepts.

For the $I_{ArcFace}$ metric, the second-best performance is achieved by \emph{W-Composite}. This can be attributed to the fact that ArcFace primarily measures identity alignment through fine-grained facial characteristics. Activating the character LoRA throughout all denoising steps promotes more consistent synthesis of these facial features, resulting in improved identity preservation.  Although the absolute $I_{ArcFace}$ scores remain relatively modest, this limitation is likely attributable to the quality of the character LoRAs themselves rather than the composition strategy. Fig.~\ref{fig:arcface_upper_bound} illustrates the behavior of $I_{ArcFace}$ when only a character LoRA is active ($N=1$), compared to the standard multi-concept generation setting ($N \geq 2$) in the \emph{ComposLoRA} benchmark. Notably, when $N=1$, all methods collapse to an identical generation process in which a single LoRA is applied with full weight at every denoising step, establishing an effective upper bound for identity alignment under the available LoRA quality. For generation with only the character LoRA activated per prompt, we obtain an $I_{ArcFace}$ score of $55.07$. As discussed above, performance degrades as the number of composed LoRAs $N$ increases. However, even at $N=5$, the relative performance drop remains limited to only $2.44\%$ for \emph{W-Switch} and $2.67\%$ for \emph{W-Composite}. This indicates that, for our methods, identity alignment remains largely preserved and is only marginally worse compared to the single-concept setting. A more detailed discussion of the limitations and trade-offs in identity preservation, as well as the effectiveness of the proposed methods, is provided in Appendix~\ref{supp_sec:identity_preservation_results}.

\begin{figure}[t]
  \centering
  \includegraphics[width=\columnwidth]{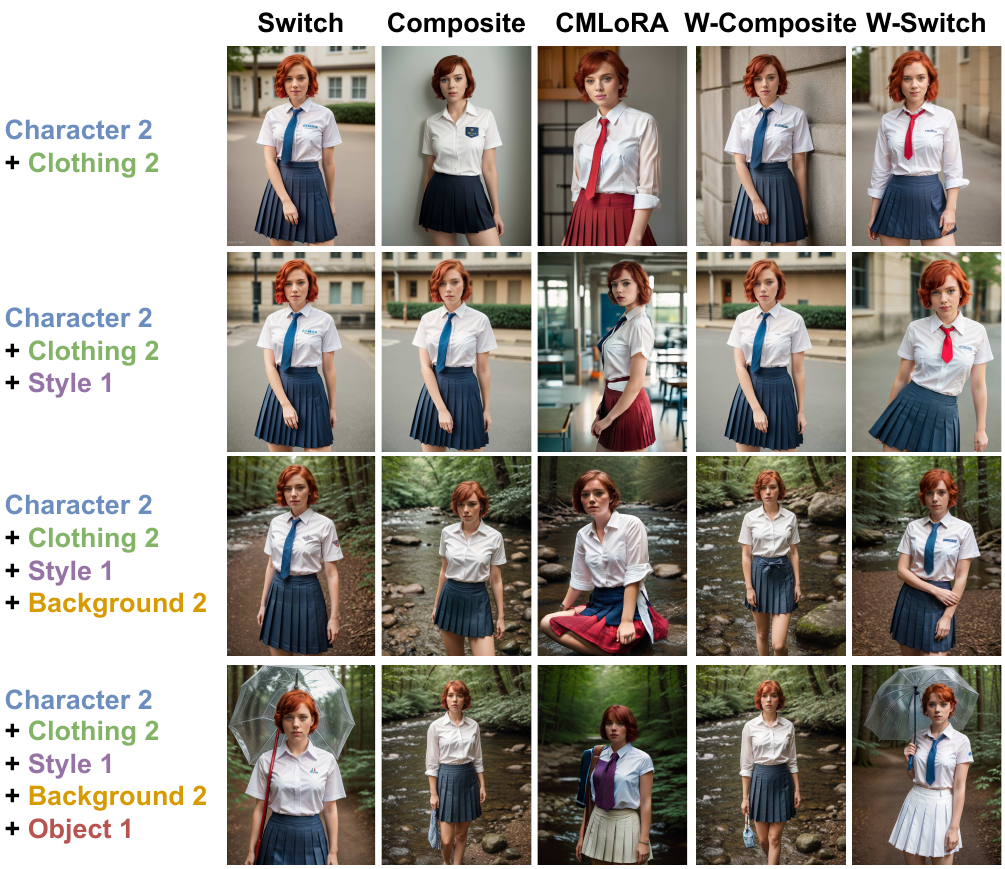}
  \caption{\textbf{Qualitative comparison of multi-LoRA composition on the \emph{ComposLoRA} testbed.} Columns show Switch~\cite{zhongmulti}, Composite~\cite{zhongmulti}, CMLoRA~\cite{zou2025cached} and the proposed \emph{W-Composite} and \emph{W-Switch} methods. Each row adds one additional LoRA.}
  \label{fig:generated_images_grid}
\end{figure}

\subsection{Qualitative Results}

Fig.~\ref{fig:generated_images_grid} presents qualitative comparisons on the \emph{ComposLoRA} benchmark across the evaluated methods as the number of composed LoRAs per prompt increases. Both \emph{W-Switch} and \emph{W-Composite} successfully preserve character identity even as the number of composed concepts grows, whereas identity degradation is pronounced for CMLoRA and becomes increasingly evident for vanilla Composite as $N$ increases. In particular, CMLoRA exhibits substantial concept interference, which leads to noticeable visual artifacts, especially in the clothing concept (third column). For compositions with fewer concepts ($N=2$ and $N=3$), \emph{W-Switch} is the only method that consistently maintains high fidelity to the original concepts as evidenced by its accurate preservation of fine-grained clothing attributes such as the blue skirt and red tie (first and second rows). Finally, for $N=5$, only Switch and \emph{W-Switch} successfully incorporate all five concepts. However, Switch often yields less natural compositions, such as incorrect placement of the umbrella shaft, whereas \emph{W-Switch} achieves a more coherent and visually plausible integration of all concepts. Additional qualitative examples are provided in Appendix~\ref{sec:qualitative_comparisons_extras}.

\paragraph*{MiniCPM Evaluation} While quantitative image-, text- and identity-alignment metrics provide useful indicators of performance, they are insufficient for capturing higher-level compositional coherence and aesthetic quality in images containing multiple concepts. Consequently, to address the limitations and potential unreliability of purely perceptual metrics, we complement our quantitative evaluation with a visual comparison of generated images using MiniCPM-V~\cite{yao2024minicpm}, following prior work~\cite{zou2025cached}. Specifically, in each evaluation round, images generated by different models using the same prompt are presented to the LLM, which assesses them along four dimensions: element integration, spatial consistency, semantic accuracy and aesthetic appeal. Additional implementation details are provided in Appendix~\ref{sec:minicpm_evaluation} and the exact evaluation prompt is reported in Fig.~\ref{fig:minicpm_prompt} in the appendix. Table~\ref{tab:minicpm_avg} reports the average scores for each evaluation dimension. Consistent with the quantitative results in Section~\ref{sec:quantitative_results}, \emph{W-Switch} achieves the best performance across all four metrics. \emph{W-Composite} also demonstrates strong results and both weighted variants consistently outperform their vanilla counterparts demonstrating the effectiveness of the prompt-aware importance weighting mechanism.

\begin{table}[t]
\centering
\caption{MiniCPM evaluation of the proposed methods against state-of-the-art baselines along four qualitative axis. Best results are shown in \textbf{bold}, second best are \underline{underlined}.}
\label{tab:minicpm_avg}
\setlength{\tabcolsep}{4pt}
\renewcommand{\arraystretch}{1.1}
\resizebox{\columnwidth}{!}{%
\begin{tabular}{lccccc}
\toprule
\multirow{2}{*}{Method}
& \multicolumn{1}{c}{Element}   & \multicolumn{1}{c}{Spatial}  & \multicolumn{1}{c}{Semantic} & \multicolumn{1}{c}{Aesthetic} & \multirow{2}{*}{Avg.} \\
& \multicolumn{1}{c}{Integration} & \multicolumn{1}{c}{Consistency} & \multicolumn{1}{c}{Accuracy} & \multicolumn{1}{c}{Appeal}    & \\
\midrule
Switch~\cite{zhongmulti}
& 8.512 & 8.418 & 8.616 & 8.142 & 8.422 \\
Composite~\cite{zhongmulti}
& 7.953 & 7.969 & 8.211 & 7.790 & 7.981 \\
CMLoRA~\cite{zou2025cached}
& 8.570 & 8.490 & 8.581 & 8.369 & 8.503 \\
\midrule
W-Composite
& \underline{8.648} & \underline{8.531} & \underline{8.637} & \underline{8.390} & \underline{8.552} \\
W-Switch
& \textbf{8.768} & \textbf{8.605} & \textbf{8.702} & \textbf{8.487} & \textbf{8.641} \\
\bottomrule
\end{tabular}%
}
\end{table}

\begin{table}[t]
\centering
\caption{User preference study results. Reported values are the fraction of trials (\%) in which each method is preferred. $\dagger$ denotes statistically significant improvement over all baseline methods.}
\label{tab:user_preference}
\setlength{\tabcolsep}{10pt}
\renewcommand{\arraystretch}{1.15}
\begin{tabular}{lc}
\toprule
Method & Preference (\%) $\uparrow$ \\
\midrule
Switch~\cite{zhongmulti} & 13.84 \\
Composite~\cite{zhongmulti} & 12.50 \\
CMLoRA~\cite{zou2025cached} & 7.59 \\
\midrule
W-Composite & 18.75 \\
W-Switch$^{\dagger}$ & \textbf{47.32} \\
\bottomrule
\end{tabular}
\end{table}

\paragraph*{Human Evaluation} Additionally, to further assess visual aesthetic quality and overall compositional coherence, we conduct a user study involving 16 human evaluators. In each evaluation round, participants are presented with a set of images generated by different models using the same target prompt and are asked to select the image that best satisfies the four criteria that were also used for the LLM-based evaluation. Table~\ref{tab:user_preference} reports the percentage of evaluation instances in which each method’s generated image was selected as the best among all candidates for the same prompt. Overall, \emph{W-Switch} is selected as the preferred method in a substantially larger fraction of cases compared to all baselines, followed by \emph{W-Composite}. Statistical significance is assessed using the Wilcoxon signed-rank test across evaluation rounds with Holm-Bonferroni correction to account for multiple comparisons confirming that \emph{W-Switch} is preferred significantly more often than all three baseline methods (\(\alpha = 0.05\)). While \emph{W-Composite} achieves higher preference rates than the baselines on average, these improvements do not reach statistical significance after correction. Additional implementation details are provided in Appendix~\ref{sec:user_study_extras}.

\subsection{Ablation Studies}

\begin{table}[t]
\centering
\caption{Ablation study on the proposed methods using different importance weighting mechanisms.}
\label{tab:paw_vs_ptw_max}
\resizebox{\columnwidth}{!}{%
\setlength{\tabcolsep}{8pt}
\renewcommand{\arraystretch}{1.15}
\begin{tabular}{llccccc}
\toprule
Method & Weights & $I_{\mathrm{CLIP}}$ & $I_{\mathrm{DINO}}$ & $I_{\mathrm{ArcFace}}$ & $T_{\mathrm{CLIP}}$ & Avg. \\
\midrule
\multirow{2}{*}{W-Composite}
 & w/ PAW & 73.29 & 49.33 & 52.73 & \textbf{35.43} & 52.70 \\
 & w/ PTW & \textbf{73.35} & \textbf{49.63} & \textbf{52.75} & 35.42 & \textbf{52.79} \\
\midrule
\multirow{2}{*}{W-Switch}
 & w/ PAW & 75.14 & 50.74 & \textbf{53.06} & \textbf{36.54} & \textbf{53.87} \\
 & w/ PTW & \textbf{75.21} & \textbf{50.88} & 52.52 & \textbf{36.54} & 53.79 \\
\bottomrule
\end{tabular}}
\end{table}

\paragraph*{PAW vs. PTW Importance Weighting Mechanisms} Table~\ref{tab:paw_vs_ptw_max} compares the two proposed importance weighting mechanisms, \emph{PAW} and \emph{PTW}, introduced in Section~\ref{weighting_mechanisms_section}, across the four evaluated quantitative metrics. Overall, \emph{PTW} yields higher $I_{CLIP}$ and $I_{DINO}$ scores, while performance in terms of $T_{CLIP}$ remains comparable across weighting strategies. The primary distinction arises for $I_{ArcFace}$, where \emph{PAW} leads to improved identity alignment for \emph{W-Switch} making it the most effective weighting mechanism on average for this method. In contrast, \emph{PTW} achieves superior $I_{ArcFace}$ performance for \emph{W-Composite} and delivers the best overall average results for this method. We note that the performance differences between the two weighting strategies are relatively small, indicating that both are viable and effective mechanisms for multi-concept LoRA composition.

\begin{table}[t]
\centering
\caption{Ablation study on reserving the final $L_{tail}$ denoising steps for character LoRAs in \emph{W-Switch}.}
\label{tab:ablation_study_reserved_steps}
\resizebox{\columnwidth}{!}{%
\setlength{\tabcolsep}{8pt}
\renewcommand{\arraystretch}{1.15}
\begin{tabular}{lccccc}
\toprule
Method & $I_{\mathrm{CLIP}}$ & $I_{\mathrm{DINO}}$ & $I_{\mathrm{ArcFace}}$ & $T_{\mathrm{CLIP}}$ & Avg. \\
\midrule
w/o $L_{tail}$ & 74.72 & \textbf{50.80} & 52.63 & 36.39 & 53.64 \\
w/ $L_{tail}$ & \textbf{75.14} & 50.74 & \textbf{53.06} & \textbf{36.54} & \textbf{53.87} \\
\bottomrule
\end{tabular}}
\end{table}

\paragraph* {Effect of $L_{tail}$} 
Table~\ref{tab:ablation_study_reserved_steps} reports the results obtained for \emph{W-Switch} with and without reserving the final $L_{tail}$ denoising steps for the character LoRA associated with the prompt. Notably, reserving these final steps results in improved average performance, yielding gains in $I_{CLIP}$, $I_{ArcFace}$, and $T_{CLIP}$ while incurring only a marginal decrease in $I_{DINO}$. This performance gain is most pronounced for $I_{ArcFace}$ which can be attributed to the emergence of high-frequency facial details during the later stages of the diffusion process~\cite{choi2022perception, park2023understanding}. Consequently, ensuring that the character LoRA remains active during these final denoising steps is particularly beneficial for preserving identity-related features.

%% file: main_sections/05_conclusions.tex
\section{Conclusions}
\label{sec:conclusions}

This paper introduces two novel methods, \emph{W-Switch} and \emph{W-Composite}, for multi-concept customization of text-to-image DMs using multiple LoRA adapters. We propose a simple yet effective, training-free importance weighting strategy that modulates the contribution of each LoRA during the denoising process. In \emph{W-Switch}, the learned weights regulate the number of denoising steps over which each LoRA is active, whereas in \emph{W-Composite}, they determine the relative influence of each LoRA on the aggregated noise prediction at every timestep. In both cases, the importance weights are derived from the semantic similarity between the target prompt embeddings and the trigger words associated with each LoRA, enabling prompt-aware multi-LoRA composition. Additionally, we introduce a novel quantitative evaluation framework that complements existing text-alignment metrics with image-based alignment and identity preservation measures. The proposed framework employs an evaluation pipeline that compares real-world reference images against automatically segmented concept regions from generated samples and is used to rigorously assess the performance of the proposed methods. Specifically, we evaluate the proposed methods on the \emph{ComposLoRA} testbed using both existing and newly introduced metrics demonstrating consistent improvements over state-of-the-art baselines. Finally, the enhanced visual quality and compositional coherence of the generated images are further validated through a MiniCPM-based evaluation and a human user preference study.

%% file: supplementary_sections/new_metrics_discussion.tex
\section{Analysis of Evaluation Metrics}
\label{sec:metrics_discussion}

\subsection{Limitations of Existing Metrics}

We identify two fundamental limitations in existing quantitative metrics for multi-concept LoRA customization of text-to-image DMs: \textit{embedding mismatch under multi-concept generation} and \textit{centroid bias induced by similarity averaging}.

\paragraph*{Embedding Mismatch under Multi-Concept Generation}
Generated images often contain multiple concepts, whereas reference images typically depict a single concept. Directly comparing their global embeddings is therefore suboptimal as the presence of additional concepts introduces noise that distorts similarity measurements. In particular, the global embedding of a multi-concept image encodes all occurring concepts and is thus expected to lie farther in the embedding latent space from the embeddings of reference images corresponding to any individual concept.

In contrast, comparing embeddings of individual concepts extracted from a multi-concept image with their respective real reference images provides a fairer and more precise evaluation. In this case, low similarity can be unambiguously attributed to poor concept fidelity in the generated image rather than to interference from other co-occurring concepts that shift the global embedding away from each concept’s reference neighborhood.

\paragraph*{Centroid Bias Induced by Similarity Averaging} 

Prior works that employ image-based alignment metrics~\cite{gu2023mix, po2024orthogonal, simsar2025loraclr} typically compute average cosine similarities between the embedding of a generated image and the embeddings of reference images corresponding to each concept. Even when the limitation of embedding mismatch between multi-concept generated images and single-concept reference images is addressed, a fundamental issue remains in such average-based metrics. Specifically, although reference images of a given concept tend to cluster in the embedding space, not every point within this neighborhood necessarily corresponds to a high-quality or realistic image that faithfully represents the concept. In contrast, a generated embedding that lies very close to a specific reference embedding is more likely to preserve the semantic fidelity of that reference instance. This issue is particularly pronounced for human reference images. Small variations in facial characteristics may induce only minor shifts in the embedding space, yet can result in generated images with noticeably degraded identity preservation.

In general, similarity measures based on averaging across reference embeddings tend to favor embeddings near the centroid of the reference set, rather than those that closely match any individual, realistic reference image. Consequently, average-based image alignment metrics may prefer embeddings that correspond to a conceptual mean which does not resemble any plausible instance of the concept. In contrast, an embedding that lies very close to a specific reference image, while being farther from the remaining references and thus receiving a lower average similarity score, may be more visually accurate and better preserve identity, particularly for human subjects.

\begin{figure}[t]
  \centering
  \includegraphics[width=0.49\textwidth]{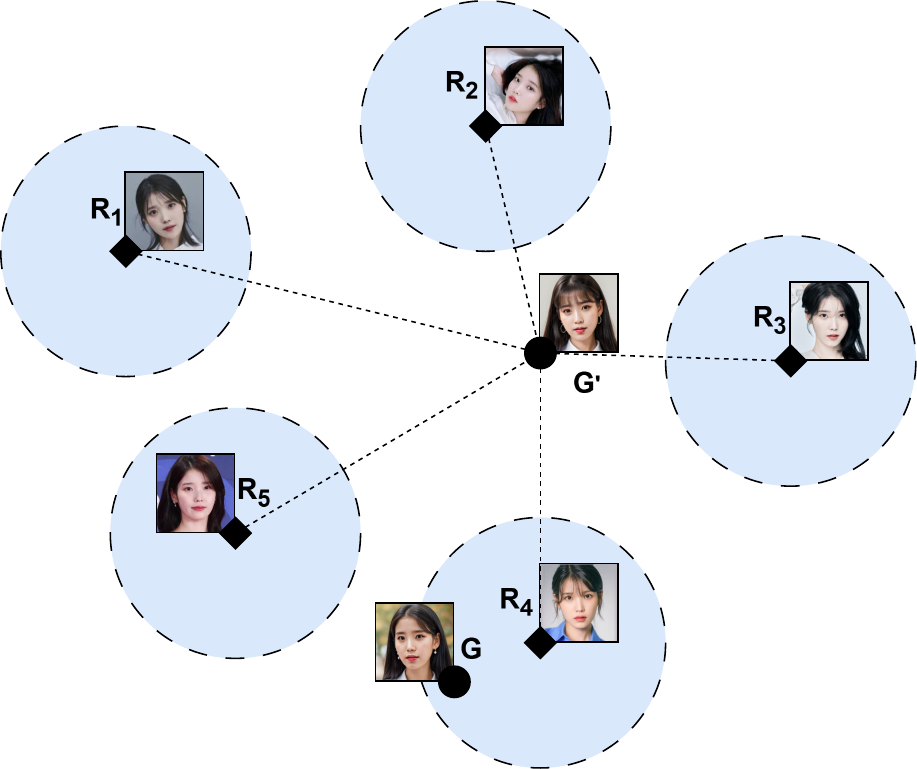}
  \caption{CLIP embedding space visualization for cropped embeddings of two generated images, $G$ and $G'$ and reference images $\{R_i\}_{i=1}^5$ of a character LoRA. Blue regions denote reference neighborhoods associated with high identity preservation and semantic fidelity.}
  \label{fig:centroid_bias}
\end{figure}

Fig.~\ref{fig:centroid_bias} illustrates this phenomenon. The set $\{R_i\}_{i=1}^5$ denotes the CLIP embeddings of the reference images associated with a specific character LoRA in the \emph{ComposLoRA} testbed while $G$ and $G'$ denote the CLIP embeddings of two generated images, cropped to include only the region corresponding to the character. The embedding $G'$ lies closer to the centroid of the reference embeddings and therefore attains the highest average similarity score ($84.07\%$). In contrast, $G$ lies in the immediate neighborhood of the reference embedding $R_4$, as illustrated by the blue circle and despite achieving a lower average cosine similarity score ($83.15\%$), exhibits higher visual fidelity and more accurately preserves the identity of the character associated with the LoRA. Consequently, we propose using the maximum cosine similarity instead of the average cosine similarity as it better rewards generated images that lie close to at least one reference embedding, thereby more faithfully preserving identity and concept fidelity.

\subsection{Single-Concept Detection and Cropping}

To address the limitation of embedding mismatch under multi-concept generation, we evaluate each concept in the image independently by first localizing the region corresponding to that concept. For character concepts, we follow the cropping implementation described in~\cite{bounareli2022finding}. Specifically, we preprocess generated images by detecting faces using the S3FD face detector~\cite{zhang2017s3fd}, followed by the estimation of 68 facial landmarks with the Face Alignment Network (FAN)~\cite{bulat2017far}. The detected face bounding box provides an initial region of interest while the FAN landmarks are used to perform a consistent, landmark-based face cropping and alignment procedure across images. Generated images in which no face is detected are discarded. 

For clothing, object and background concepts, we automate the cropping process using SAM3~\cite{carion2025sam}, which enables object detection and segmentation via short textual prompts. We distinguish between foreground and background concepts, where foreground concepts include characters, clothing and objects, and background concepts correspond to the scene background. For each foreground concept, we provide SAM3 with the trigger words associated with the corresponding LoRA as the text prompt. We then extract the resulting segmentation mask and crop the image to the minimal rectangular region enclosing the mask. As portions of the background may still be visible within this crop, we further reduce background influence on the embedding by replacing background pixels with a blurred (local-mean) version.

\begin{figure}[t]
    \centering
    \includegraphics[width=\columnwidth]{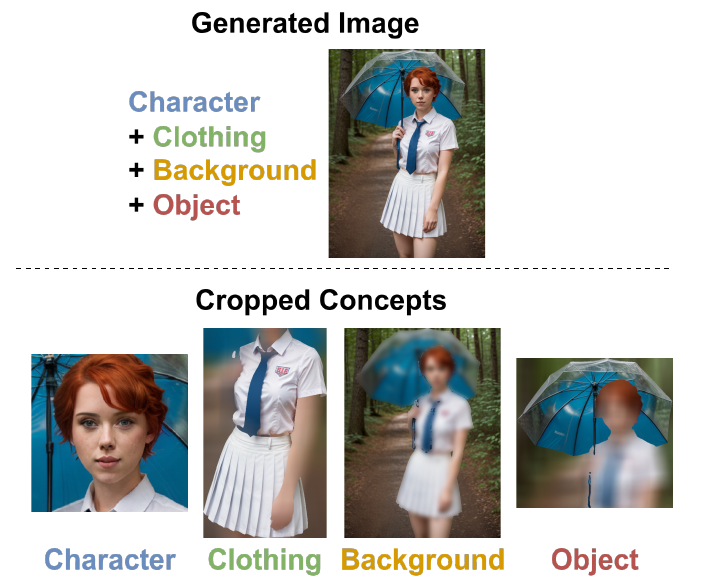}
    \caption{Example of concept-specific cropped regions extracted from a multi-concept image generated using four concept LoRAs.}
    \label{fig:cropped_concepts}
\end{figure}

In contrast, background concepts typically span the entire image, with foreground objects interleaved throughout, which can skew the resulting embeddings. To obtain embeddings that primarily capture background semantics while minimizing foreground influence, we apply the complementary procedure to that used for foreground concepts. Specifically, we use SAM3 to detect and segment all foreground concepts in the image, including characters. As characters are highly instance-specific and SAM3 is more effective with broad object categories than with individual identities that may not have been observed during training, we employ generic textual prompts such as “a man” or “a woman” to obtain character segmentation masks. After extracting all foreground masks, we reduce their influence by replacing pixels within these masks with a blurred (local-mean) version. Finally, style LoRAs act as global filters that affect the entire image and as a result no cropping is applied for these concepts. Fig.~\ref{fig:cropped_concepts} illustrates examples of cropped foreground and background regions extracted from a multi-concept generated image.

\subsection{Unified Evaluation Pipeline}

\begin{figure*}[t]
    \centering
    \includegraphics[width=0.9\textwidth]{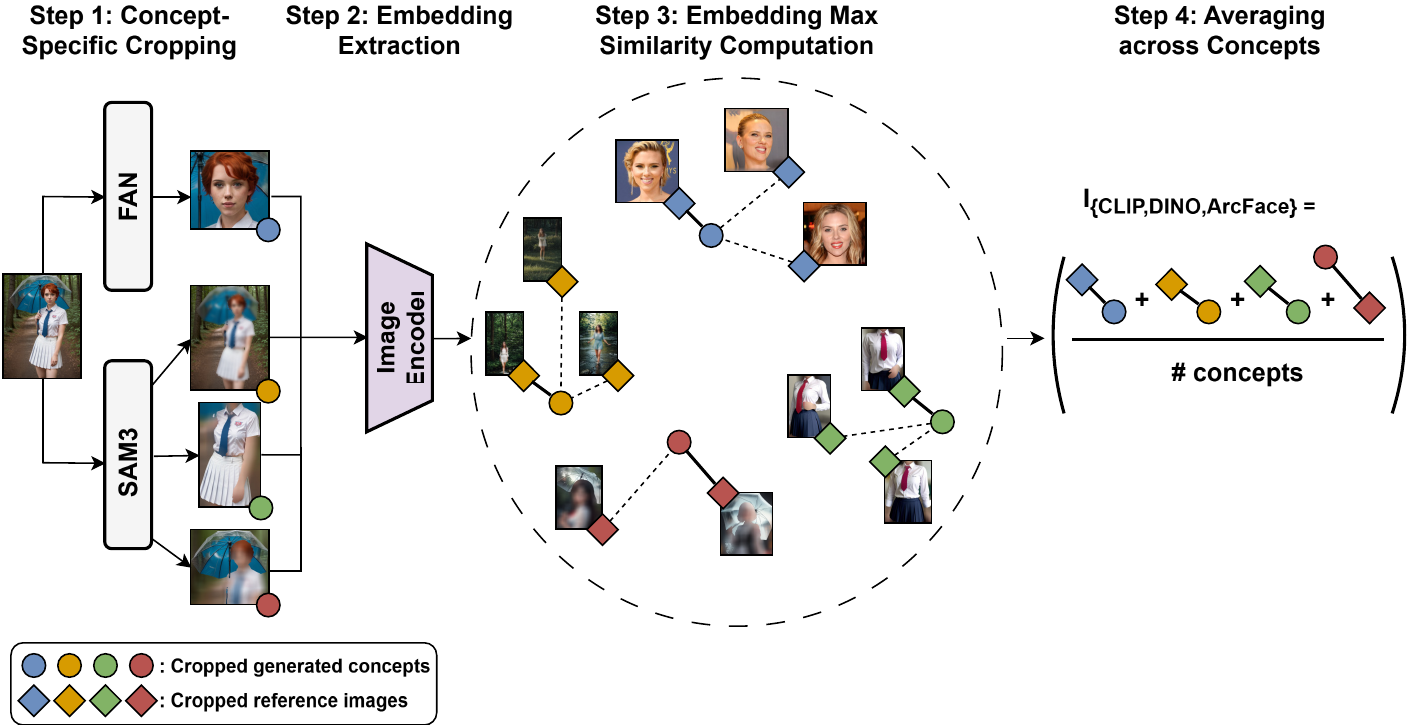}
    \caption{Overview of the proposed evaluation pipeline for $I_{CLIP}$, $I_{DINO}$ and $I_{ArcFace}$ based on concept-specific cropping, encoder-based,  max-pooled similarity aggregation and averaging across concepts.}
    \label{fig:full_evaluation_pipeline}
\end{figure*}

Fig.~\ref{fig:full_evaluation_pipeline} illustrates the complete evaluation pipeline used to compute the proposed image-based alignment and identity preservation metrics, $I_{CLIP}$, $I_{DINO}$ and $I_{ArcFace}$. Given a generated image composed of multiple LoRAs the pipeline proceeds through four stages: concept localization, concept-specific cropping, embedding extraction and similarity aggregation.

First, individual concepts present in the generated image are localized using automated detectors and cropped into separate image regions. Second, each concept-specific crop is independently embedded using a fixed image encoder (CLIP, DINOv2, ArcFace) depending on the metric being computed. Next, for each concept, similarities are computed between the embedding of the cropped region and the embeddings of its corresponding real reference images. Finally, instead of averaging similarities across reference images, we employ a max-pooling strategy that selects the maximum cosine similarity for each concept, followed by averaging across concepts.

%% file: supplementary_sections/experimental_results_extras.tex
\section{Additional Experimental Results}
\label{sec:experimental_results_extras}

\begin{figure*}[t]
    \centering
    \includegraphics[width=0.85\textwidth]{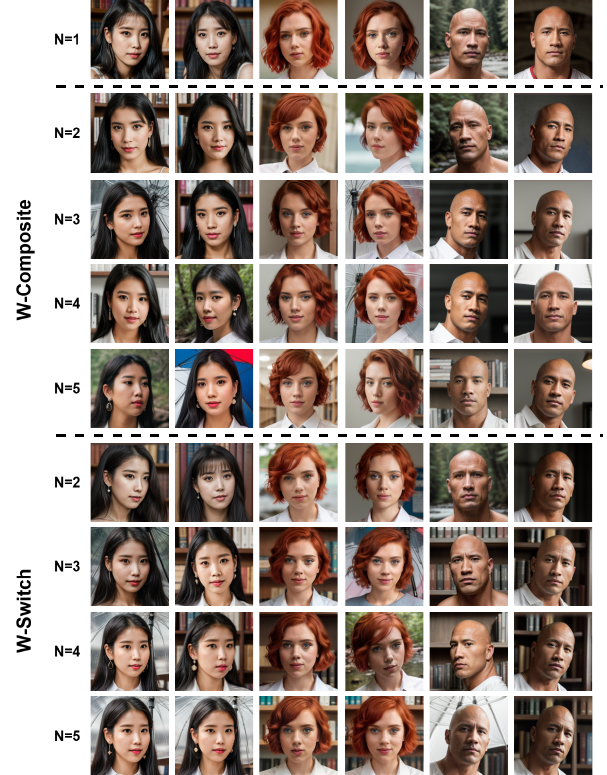}
    \caption{Generated images for the three character LoRAs using \emph{W-Composite} and \emph{W-Switch} across increasing numbers of composed LoRAs $N$, compared against the single-LoRA generation setting ($N{=}1$).}
    \label{fig:arcface_grid_supplementary}
\end{figure*}

\begin{table*}[t]
\centering
\caption{$I_{\text{ArcFace}}$ metric values and relative degradation ($\Delta I_{\text{ArcFace}}$) measured with respect to the character-only LoRA baseline as the number of composed LoRAs increases ($N{=}2$–$5$) in the \emph{ComposLoRA} testbed.}
\label{tab:arcface_degradation}
\setlength{\tabcolsep}{6pt}
\renewcommand{\arraystretch}{1.15}
\resizebox{\textwidth}{!}{%
\begin{tabular}{lcccccc}
\toprule
\multirow{2}{*}{Method}
& \multicolumn{1}{c}{N=1}
& \multicolumn{1}{c}{N=2}
& \multicolumn{1}{c}{N=3}
& \multicolumn{1}{c}{N=4}
& \multicolumn{1}{c}{N=5}
& \multicolumn{1}{c}{Avg.} \\
\cmidrule(lr){2-2}
\cmidrule(lr){3-3}
\cmidrule(lr){4-4}
\cmidrule(lr){5-5}
\cmidrule(lr){6-6}
\cmidrule(lr){7-7}
& $I_{ArcFace}$
& $I_{ArcFace}\rightarrow \Delta I_{ArcFace}$
& $I_{ArcFace}\rightarrow \Delta I_{ArcFace}$
& $I_{ArcFace}\rightarrow \Delta I_{ArcFace}$
& $I_{ArcFace}\rightarrow \Delta I_{ArcFace}$
& $I_{ArcFace} \rightarrow \Delta I_{ArcFace}$ \\
\midrule
Switch~\cite{zhongmulti}
& 55.07
& 51.80 $\rightarrow$ -3.27
& 52.02 $\rightarrow$ -3.05
& 50.22 $\rightarrow$ -4.85
& 50.99 $\rightarrow$ -4.17
& 51.24 $\rightarrow$ -3.83 \\
Composite~\cite{zhongmulti}
& 55.07
& \textbf{55.79} $\rightarrow$ \textbf{+0.72}
& 51.46 $\rightarrow$ -3.61
& 51.80 $\rightarrow$ -3.27
& 50.74 $\rightarrow$ -4.33
& 52.45 $\rightarrow$ -2.62 \\
CMLoRA~\cite{zou2025cached}
& 55.07
& 52.32 $\rightarrow$ -2.75
& 50.46 $\rightarrow$ -4.61
& 48.47 $\rightarrow$ -6.60
& 48.85 $\rightarrow$ -6.22
& 50.03 $\rightarrow$ -5.04 \\
\midrule
W-Composite
& 55.07
& 54.11 $\rightarrow$ -0.96
& 52.42 $\rightarrow$ -2.65
& 52.40 $\rightarrow$ -2.67
& 52.05 $\rightarrow$ -3.02
& 52.75 $\rightarrow$ -2.32 \\
W-Switch
& 55.07
& 53.57 $\rightarrow$ -1.50
& \textbf{53.34} $\rightarrow$ \textbf{-1.73}
& \textbf{52.63} $\rightarrow$ \textbf{-2.44}
& \textbf{52.68} $\rightarrow$ \textbf{-2.01}
& \textbf{53.06} $\rightarrow$ \textbf{-2.01} \\
\bottomrule
\end{tabular}
}
\end{table*}

\begin{table*}[t]
\centering
\caption{Full MiniCPM evaluation results for an increasing number of composed LoRAs in the \emph{ComposLoRA} testbed. Best results are shown in \textbf{bold}, second best are \underline{underlined}.}
\label{tab:full_minicpm}
\setlength{\tabcolsep}{4pt}
\renewcommand{\arraystretch}{1.15}
\resizebox{\textwidth}{!}{%
\begin{tabular}{lcccccccccccccccccccccc}
\toprule
\multirow{2}{*}{Method}
 & \multicolumn{5}{c}{Element Integration}
 & \multicolumn{5}{c}{Spatial Consistency}
 & \multicolumn{5}{c}{Semantic Accuracy}
 & \multicolumn{5}{c}{Aesthetic Appeal}
 & \multirow{2}{*}{Avg.} \\
\cmidrule(lr){2-6}
\cmidrule(lr){7-11}
\cmidrule(lr){12-16}
\cmidrule(lr){17-21}
 & N=2 & N=3 & N=4 & N=5 & Avg.
 & N=2 & N=3 & N=4 & N=5 & Avg.
 & N=2 & N=3 & N=4 & N=5 & Avg.
 & N=2 & N=3 & N=4 & N=5 & Avg.
 &  \\
\midrule
Switch~\cite{zhongmulti}
 & 8.434 & 8.529 & 8.543 & 8.540 & 8.512
 & 8.355 & 8.437 & 8.426 & 8.452 & 8.418
 & 8.526 & \underline{8.621} & \underline{8.721} & 8.597 & 8.616
 & 8.132 & 8.160 & 8.161 & 8.116 & 8.142
 & 8.422 \\
Composite~\cite{zhongmulti}
 & 7.842 & 7.975 & 8.010 & 7.984 & 7.953
 & 7.763 & 8.059 & 8.054 & 8.000 & 7.969
 & 8.105 & 8.162 & 8.310 & 8.266 & 8.211
 & 7.724 & 7.789 & 7.822 & 7.823 & 7.790
 & 7.981 \\
CMLoRA~\cite{zou2025cached}
 & 8.632 & 8.591 & 8.500 & 8.556 & 8.570
 & \textbf{8.605} & 8.495 & 8.400 & 8.460 & 8.490
 & 8.632 & 8.577 & 8.540 & 8.573 & 8.581
 & \textbf{8.526} & 8.327 & 8.259 & 8.363 & 8.369
 & 8.503 \\
\midrule
W-Composite
 & \underline{8.645} & \underline{8.686} & \underline{8.648} & \underline{8.613} & \underline{8.648}
 & \underline{8.566} & \underline{8.519} & \underline{8.492} & \underline{8.548} & \underline{8.531}
 & \underline{8.645} & \textbf{8.624} & 8.650 & \underline{8.629} & \underline{8.637}
 & \underline{8.474} & \underline{8.357} & \underline{8.372} & 8.355 & \underline{8.390}
 & \underline{8.552} \\
W-Switch
 & \textbf{8.671} & \textbf{8.743} & \textbf{8.819} & \textbf{8.839} & \textbf{8.768}
 & 8.553 & \textbf{8.552} & \textbf{8.612} & \textbf{8.702} & \textbf{8.605}
 & \textbf{8.658} & 8.619 & \textbf{8.771} & \textbf{8.758} & \textbf{8.702}
 & 8.461 & \textbf{8.419} & \textbf{8.543} & \textbf{8.524} & \textbf{8.487}
 & \textbf{8.641} \\
\bottomrule
\end{tabular}%
}
\end{table*}

\begin{table*}[t]
\centering
\caption{Full ablation study results on \emph{W-Switch} and \emph{W-Composite} using different importance weighting mechanisms under different number of LoRAs in the \emph{ComposLoRA} testbed.}
\label{tab:paw_ptw_ablation_supplementary}
\setlength{\tabcolsep}{3pt}
\renewcommand{\arraystretch}{1.15}
\resizebox{\textwidth}{!}{%
\begin{tabular}{lcccccccccccccccccccccc}
\toprule
\multirow{2}{*}{Method}
 & \multicolumn{5}{c}{$I_{\mathrm{CLIP}}$}
 & \multicolumn{5}{c}{$I_{\mathrm{CLIP}}$}
 & \multicolumn{5}{c}{$I_{\mathrm{ArcFace}}$}
 & \multicolumn{5}{c}{$T_{\mathrm{CLIP}}$}
 & \multirow{2}{*}{Avg.} \\
\cmidrule(lr){2-6}
\cmidrule(lr){7-11}
\cmidrule(lr){12-16}
\cmidrule(lr){17-21}
 & N=2 & N=3 & N=4 & N=5 & Avg.
 & N=2 & N=3 & N=4 & N=5 & Avg.
 & N=2 & N=3 & N=4 & N=5 & Avg.
 & N=2 & N=3 & N=4 & N=5 & Avg.
 &  \\
\midrule
W-Composite w/ PAW
 & 76.07 & 74.27 & \textbf{72.55} & 70.28 & 73.29
 & 52.83 & 50.80 & 48.47 & \textbf{45.20} & 49.33
 & 53.67 & \textbf{52.76} & \textbf{52.49} & 51.98 & 52.73
 & 34.60 & \textbf{35.59} & \textbf{35.69} & \textbf{35.83} & \textbf{35.43} & 52.70 \\
W-Composite w/ PTW
 & \textbf{76.20} & \textbf{74.34} & \textbf{72.55} & \textbf{70.30} & \textbf{73.35}	
 & \textbf{53.70} & \textbf{51.16} & \textbf{48.66} & 44.98 & \textbf{49.63}	
 & \textbf{54.11} & 52.42 & 52.40 & \textbf{52.05} & \textbf{52.75}
 & \textbf{34.70} & \textbf{35.59} & \textbf{35.69} & 35.71	& 35.42 & \textbf{52.79}\\
 \midrule
W-Switch w/ PAW
 & 76.23 & \textbf{75.80} & \textbf{74.72} & 73.80 & 75.14
 & 54.20 & 51.57 & \textbf{49.14} & \textbf{48.03} & 50.74
 & \textbf{53.57} & \textbf{53.34} & \textbf{52.63} & \textbf{52.68} & \textbf{53.06}
 & 34.78 & \textbf{36.70} & \textbf{37.12} & 37.55 & \textbf{36.54} & \textbf{53.87} \\
W-Switch w/ PTW
 & \textbf{76.55} & 75.78 & 74.59 & \textbf{73.91} & \textbf{75.21}
 & \textbf{54.61} & \textbf{51.94} & 49.00  & 47.98 & \textbf{50.88}
 & 52.37 & 53.26 & 52.27 & 52.16 & 52.52 
 & \textbf{34.99} & 36.56 & 37.00 & \textbf{37.60} & \textbf{36.54} & 53.79 \\
\bottomrule
\end{tabular}%
}
\end{table*}

\begin{table*}[t]
\centering
\caption{Full ablation study results on reserving the final $L_{tail}$ steps for character LoRAs in \emph{W-Switch} in the \emph{ComposLoRA} testbed.}
\label{tab:ltail_ablation_supplementary}
\setlength{\tabcolsep}{3pt}
\renewcommand{\arraystretch}{1.15}
\resizebox{\textwidth}{!}{%
\begin{tabular}{lcccccccccccccccccccccc}
\toprule
\multirow{2}{*}{Method}
 & \multicolumn{5}{c}{$I_{\mathrm{CLIP}}$}
 & \multicolumn{5}{c}{$I_{\mathrm{DINO}}$}
 & \multicolumn{5}{c}{$I_{\mathrm{ArcFace}}$}
 & \multicolumn{5}{c}{$T_{\mathrm{CLIP}}$}
 & \multirow{2}{*}{Avg.} \\
\cmidrule(lr){2-6}
\cmidrule(lr){7-11}
\cmidrule(lr){12-16}
\cmidrule(lr){17-21}
 & N=2 & N=3 & N=4 & N=5 & Avg.
 & N=2 & N=3 & N=4 & N=5 & Avg.
 & N=2 & N=3 & N=4 & N=5 & Avg.
 & N=2 & N=3 & N=4 & N=5 & Avg.
 &  \\
\midrule
W-Switch w/o $L_{tail}$
 & 75.95 & 75.23 & 74.12 & 73.56 & 74.72 
 & 54.05 & \textbf{51.86} & \textbf{49.19} & \textbf{48.11} & \textbf{50.80}
 & 53.10 & 52.64 & 51.97 & \textbf{52.80} & 52.63
 & \textbf{34.81} & 36.21 & 37.01 & 37.52 & 36.39 & 53.64 \\
W-Switch w/ $L_{tail}$
 & \textbf{76.23} & \textbf{75.80} & \textbf{74.72} & \textbf{73.80} & \textbf{75.14}
 & \textbf{54.20} & 51.57 & 49.14 & 48.03 & 50.74
 & \textbf{53.57} & \textbf{53.34} & \textbf{52.63} & 52.68 & \textbf{53.06}
 & 34.78 & \textbf{36.70} & \textbf{37.12} & \textbf{37.55} & \textbf{36.54} & \textbf{53.87} \\
\bottomrule
\end{tabular}%
}
\end{table*}

\subsection{Identity Preservation Results} \label{supp_sec:identity_preservation_results}

Table~\ref{tab:arcface_degradation} reports the identity preservation degradation incurred when transitioning from the single-LoRA generation setting to multi-LoRA composition. Specifically, we measure the change in $I_{ArcFace}$ as the number of composed LoRAs $N$ increases on the \emph{ComposLoRA} testbed, relative to the baseline performance obtained when only the character LoRA corresponding to the prompt is activated. This degradation is quantified using $\Delta I_{ArcFace,N}$, defined as:

\begin{equation}
\Delta I_{ArcFace,N} = I_{ArcFace,N} - I_{ArcFace,1},
\end{equation}

where the subscript $N$ denotes the number of activated LoRAs used to compute the metric. Notably, the $N{=}1$ setting serves as an upper bound in most cases, representing the best attainable identity preservation. As the number of composed LoRAs increases, identity alignment is expected to degrade since combining multiple LoRAs can introduce concept interference, even when only a single character is present in the generated image.

While $I_{ArcFace}$ is a useful metric for assessing identity preservation in multi-concept image composition, it is important to interpret this metric in conjunction with the performance of the character LoRAs in the single-LoRA setting where no additional LoRAs are activated. If a character LoRA produces suboptimal identity preservation in isolation, it is unreasonable to expect improved performance in the more challenging multi-LoRA composition setting. Accordingly, we evaluate both the absolute performance of character LoRAs under single-LoRA generation and the degree of identity degradation incurred when transitioning to multi-LoRA composition, which we quantify using $\Delta I_{ArcFace}$. As the number of composed concepts increases, the three examined baseline methods exhibit a noticeable decline in $I_{ArcFace}$, on the order of 3–6\%. In contrast, although \emph{W-Switch} and \emph{W-Composite} also experience some degradation, it remains limited to approximately 2–3\%. This suggests that the comparatively lower absolute $I_{ArcFace}$ values observed for these methods are primarily attributable to the inherent quality of the underlying character LoRAs, rather than to increased identity interference during multi-LoRA composition, as evidenced by the minimal performance drop when additional LoRAs are introduced. Finally, Fig.~\ref{fig:arcface_grid_supplementary} presents representative examples of generated images for $N{=}1$ and $N{\geq}2$ using \emph{W-Switch} and \emph{W-Composite} across the three examined character LoRAs. Facial regions are cropped using the same FAN face detector employed in the computation of $I_{ArcFace}$. The results illustrate that while identity fidelity is not consistently high, the observed limitations primarily originate from the inherent quality of the underlying character LoRAs rather than from increased interference as additional concepts are composed.

\subsection{MiniCPM Evaluation Results}

In addition to the results reported in Table~\ref{tab:minicpm_avg} of the main paper, we provide the complete MiniCPM-based evaluation results on the \emph{ComposLoRA} testbed for an increasing number of composed LoRAs in Table~\ref{tab:full_minicpm}. Overall, \emph{W-Switch} achieves the strongest performance in the majority of cases, followed by \emph{W-Composite}. These results demonstrate that the proposed methods excel in perceptual qualities that are not well captured by standard quantitative metrics, including overall aesthetic quality and harmonious multi-concept integration. Notably, as the number of composed LoRAs increases, \emph{W-Switch} increasingly dominates the results, underscoring its robustness in more challenging multi-concept composition settings. 

%% file: supplementary_sections/ablation_study_extras.tex
\section{Additional Ablation Study Results}
\label{sec:ablation_study_extras}

\subsection{PAW vs. PTW Importance Weighting Mechanisms}

Beyond the results in Table~\ref{tab:paw_vs_ptw_max} of the main paper, Table~\ref{tab:paw_ptw_ablation_supplementary} reports the full ablation results for \emph{W-Composite} and \emph{W-Switch} with $N=2$–$5$ composed LoRAs under the \emph{PAW} and \emph{PTW} weighting schemes. For \emph{W-Composite}, \emph{PTW} generally outperforms \emph{PAW}, whereas for \emph{W-Switch} the two weighting schemes yield comparable performance across $I_{CLIP}$, $I_{DINO}$ and $T_{CLIP}$. However, \emph{PAW} consistently achieves higher performance for \emph{W-Switch} across all $N$, leading to superior average results.

\subsection{Effect of $L_{tail}$}

Beyond Table~\ref{tab:ablation_study_reserved_steps} of the main paper, Table~\ref{tab:ltail_ablation_supplementary} reports full ablation results for \emph{W-Switch} with and without reserving the final $L_{tail}$ denoising steps for character LoRAs, evaluated on the \emph{ComposLoRA} testbed for $N=2$–$5$. Implementing \emph{W-Switch} with reserved $L_{tail}$ denoising steps consistently improves performance across $I_{CLIP}$, $I_{ArcFace}$ and $T_{CLIP}$ while incurring only a minor average degradation in $I_{DINO}$, confirming the effectiveness of this simple modification in enhancing overall performance.

%% file: supplementary_sections/mllm_evaluation.tex
\section{MLLM-based Evaluation with MiniCPM} \label{sec:minicpm_evaluation}

Existing quantitative evaluation metrics for multi-subject composition in text-to-image generation have primarily focused on text–image alignment (e.g., CLIPScore~\cite{hessel2021clipscore}), alignment between generated images and real reference images (e.g., $I_{CLIP}$ and $I_{DINO}$~\cite{gal2022image}) and identity preservation (e.g., $I_{ArcFace}$~\cite{deng2019arcface}). While these metrics provide an intuitive means of comparing different methods, they are inherently limited in capturing qualities that are central to multi-subject composition. Specifically, they focus on individual concepts in isolation rather than on the joint composition of multiple concepts within a single image. As a result, they fail to adequately assess aspects such as the seamless integration of concepts, spatial consistency, preservation of semantic fidelity and the overall aesthetic quality of the generated image.

In recent years, an increasing body of work has leveraged multimodal LLMs, such as GPT-4V and MiniCPM-V~\cite{yao2024minicpm}, as evaluators of these more abstract qualities in generated images, capitalizing on their strong multimodal reasoning capabilities~\cite{zhang2023gpt}. Following prior work~\cite{zou2025cached}, we adopt MiniCPM-V to compare images produced by state-of-the-art baseline methods against those generated by our proposed approaches. Unlike prior approaches that rely on pairwise comparisons~\cite{foteinopoulou2025loratorio, zhongmulti, zou2025cached}, we adopt a multi-way comparison setting in which images from each model are presented simultaneously and evaluated jointly. We find that this formulation facilitates more direct and consistent comparisons across methods, as all models are assessed within a shared reference framework for score assignment. We conduct this evaluation using all images generated on the \emph{ComposLoRA} testbed while ensuring that images compared across methods are produced using identical prompts and random seeds. We employ a blind evaluation protocol, in which the evaluator scores the images according to predefined criteria without access to the identity of the generating model. Specifically, we assess four dimensions: element integration, spatial consistency, semantic accuracy, and aesthetic quality. Each criterion is rated on a scale from 0 to 10, with higher scores indicating superior performance. The evaluation prompt provided to the model is shown in Fig.~\ref{fig:minicpm_prompt}.

\begin{figure*}[t]
\centering
\begin{promptbox}{MiniCPM-V Evaluation Prompt}
\begin{lstlisting}[style=promptstyle]
<IMAGE_1> <IMAGE_2> <IMAGE_3> <IMAGE_4> <IMAGE_5> <IMAGE_6>

You are given 6 images generated by different text-to-image methods. The expected concepts in the image include: <PROMPT>

Key attributes:

1) Element Integration: How seamlessly different elements are combined within the image.
Criteria:
- Visual Cohesion: Evaluate whether the elements appear as part of a unified scene, rather than as disjointed parts.
- Object Overlap and Interaction: Check for natural overlaps and interactions between objects, ensuring no awkward placements or intersections.

2) Spatial Consistency: Uniformity in style, lighting, and perspective across all elements.
Criteria:
- Stylistic Uniformity: Ensure that all elements share a consistent artistic style (e.g., realism, cartoonish).
- Lighting and Shadows: Verify that light sources and shadow directions are consistent, contributing to a realistic portrayal.
- Perspective Alignment: Confirm that elements adhere to a shared perspective, with no mismatched viewpoints.

3) Semantic Accuracy: Correct interpretation and representation of each element as described in the prompt.
Criteria:
- Object Accuracy: Objects should align with their descriptions in terms of type, attributes, and context.
- Action and Interaction: Actions or interactions between objects should be depicted accurately and appropriately.

4) Aesthetic Quality: Overall visual appeal and artistic quality of the generated image.
Criteria:
- Color Harmony: The use of color palettes should be visually pleasing and fitting for the scene.
- Composition Balance: Elements should be arranged in a balanced way to create an engaging and harmonious composition.
- Clarity and Sharpness: The image should be clear, with well-defined elements, free from unwanted blurriness or distortion.

Please rate each image independently on: Integration of Elements, Consistency in Composition, Accuracy in Depiction, Visual Appeal.

Respond ONLY with a JSON array where each entry has: "method", "integration", "consistency", "accuracy", "appeal". Use numeric scores out of 10. The numbers can be any value between 0 and 10, and you may use both integer and .5 scores, for example: 5, 5.5, 7.5. No extra text.
\end{lstlisting}
\end{promptbox}
\caption{Prompt used for the MiniCPM-based image quality evaluation.}
\label{fig:minicpm_prompt}
\end{figure*}

\begin{figure}[t]
  \centering
  \includegraphics[width=\columnwidth]{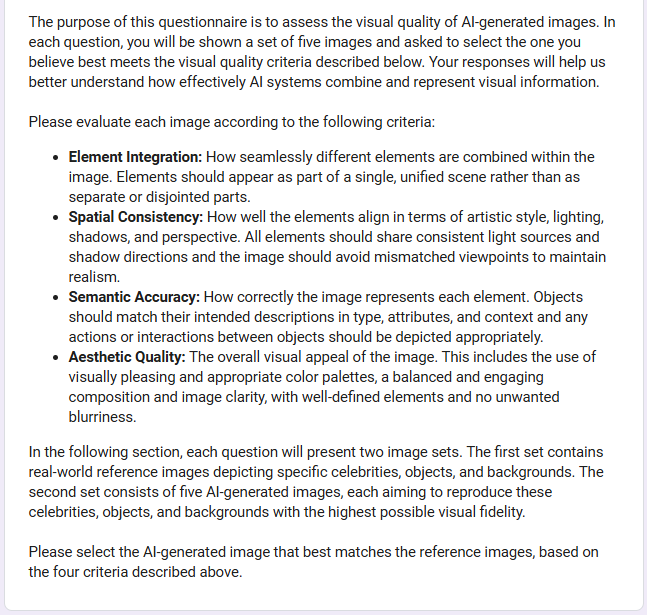}
  \caption{Instructions shown to the user study participants outlining the evaluation procedure and assessment criteria.}
  \label{fig:user_study_instructions}
\end{figure}

\begin{figure}[t]
  \centering
  \includegraphics[width=\columnwidth]{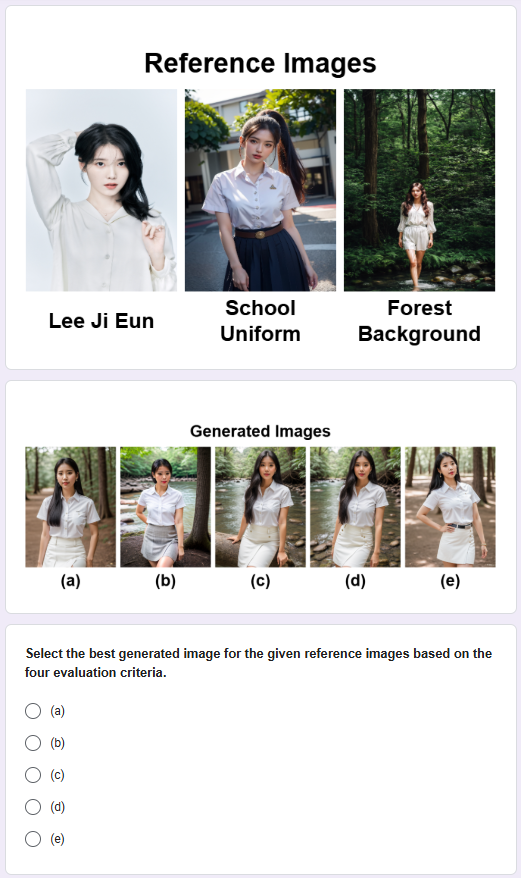}
  \caption{Sample evaluation question used in the human preference study.}
  \label{fig:user_study_question}
\end{figure}

%% file: supplementary_sections/user_study_extras.tex
\section{User Study Evaluation}
\label{sec:user_study_extras}

For the human-based evaluation of the proposed methods, we conducted a user study involving responses from 16 participants. We constructed 14 sets of concept combinations, each comprising between two and five concepts. For each combination, we generated one image using each of the two proposed methods as well as the three state-of-the-art baseline models considered in this paper, ensuring that all methods used the same random seed for fair comparison. In each evaluation question, participants were shown a set of reference images corresponding to the individual concepts, together with the five generated images produced by the anonymized methods. Participants were asked to select the generated image that best matched the reference images according to the four criteria (element integration, spatial consistency, semantic accuracy, aesthetic quality) which were also employed in the LLM-based evaluation using MiniCPM, as presented in Section~\ref{sec:minicpm_evaluation}. Fig.~\ref{fig:user_study_instructions} presents the instructions provided to the participants while Fig.~\ref{fig:user_study_question} illustrates an example of the evaluation questions shown during the study.

Finally, regarding the statistical analysis of the human preference study, each evaluation round is treated as a paired observation and for a given round, we compare the number of participant selections assigned to two methods under identical prompting conditions resulting in 14 paired samples per comparison. Statistical significance is assessed using a two-sided Wilcoxon signed-rank test applied across evaluation rounds. Pairwise comparisons are conducted between each proposed method (\emph{W-Switch} and \emph{W-Composite}) and all three baseline methods, yielding a total of six statistical tests. To account for multiple comparisons, Holm-Bonferroni correction is applied with a family-wise significance level of $\alpha = 0.05$. For completeness, Table~\ref{tab:user_study_stats} reports the raw p-values obtained from the Wilcoxon signed-rank tests prior to correction. Consistent with the results discussed in the main paper, \emph{W-Switch} demonstrates statistically significant improvements over all baseline methods after correction, whereas the improvements of \emph{W-Composite} do not reach statistical significance.

\begin{table}[t]
\centering
\caption{Raw p-values from the Wilcoxon signed-rank test for the human preference study. Holm--Bonferroni correction is applied across all six comparisons ($\alpha=0.05$).}
\label{tab:user_study_stats}
\setlength{\tabcolsep}{6pt}
\renewcommand{\arraystretch}{1.1}
\begin{tabular}{lc}
\toprule
Comparison & Raw $p$-value \\
\midrule
W-Composite vs Switch & 0.4631 \\
W-Composite vs Composite & 0.4263 \\
W-Composite vs CMLoRA & 0.0278 \\
\midrule
W-Switch vs Switch & 0.0001 \\
W-Switch vs Composite & 0.0014 \\
W-Switch vs CMLoRA & 0.0002 \\
\bottomrule
\end{tabular}
\end{table}

\begin{figure}[t]
  \centering
  \includegraphics[width=\columnwidth]{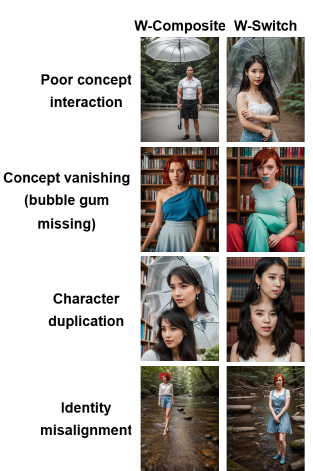}
  \caption{Examples of failure cases for the two proposed methods.}
  \label{fig:failure_cases}
\end{figure}

%% file: supplementary_sections/limitations_error_cases.tex
\section{Limitations and Error Cases}
\label{sec:limitations_error_cases}

The first limitation concerns the lack of fine-grained control over specific image regions during the generation process. While the proposed methods introduce additional flexibility through importance weighting mechanisms that modulate the contribution of different concepts, this control remains global rather than spatially localized. In particular, emphasis can be adjusted either by assigning relative weights in the weighted aggregation of \emph{W-Composite} or by varying the number of denoising steps during which a concept-specific LoRA is activated in \emph{W-Switch}. However, none of these mechanisms explicitly enables region-level control within the generated image. This limitation arises because the proposed methods are training-free and do not leverage prior spatial information, such as region- or layout-level constraints, including bounding boxes or masked attention maps. While this design choice makes the methods computationally efficient, lightweight and easy to deploy, it also restricts their ability to explicitly model and enforce complex spatial relationships within the generated images. Fig.~\ref{fig:failure_cases} illustrates representative failure cases of our approaches. In particular, the absence of regionally controllable sampling can result in poor spatial relationships between concepts, as shown in the first row of Fig.~\ref{fig:failure_cases}, where the interaction between the character and the umbrella object is not properly realized. Moreover, the lack of explicit mechanisms for enforcing spatial localization may give rise to semantic inconsistencies including concept vanishing (second row of Fig.~\ref{fig:failure_cases}) and unintended character duplication (third row of Fig.~\ref{fig:failure_cases}).

Our method implicitly relies on the assumption that individual LoRA adapters are trained using semantically consistent and well-curated datasets. In real-world settings, however, the quality of such adapters can vary widely, especially when obtained from community-driven repositories such as CivitAI, where training data are often undocumented or lack standardization. Furthermore, LoRA adapters employed at inference time are often heterogeneous with respect to their optimal scaling and hyperparameter configurations. Applying a uniform treatment across such adapters may inadvertently favor those with stronger or more aggressive activations, thereby introducing bias in the composed output.

Similarly, we observe that image quality is influenced by the choice of base model. A notable limitation of commonly used base models concerns the generation of small faces. In the case of SD, information loss introduced by the VAE can degrade the quality of full-body character synthesis, particularly in regions containing small facial features, leading to diminished facial detail, as illustrated in the last row of Fig.~\ref{fig:failure_cases}.

Finally, since all LoRA modules used in our experiments are sourced from CivitAI and lack publicly available training details, our results should be interpreted with this uncertainty in mind. This limitation is particularly relevant for identity preservation. As discussed in Section~\ref{supp_sec:identity_preservation_results}, even single-concept generation using only character LoRAs yields performance that is only marginally higher than that achieved by our proposed multi-concept customization methods.

%% file: supplementary_sections/future_work.tex
\section{Future Work}
\label{sec:future_work}

Several promising directions remain for future exploration. First, extending the proposed framework to 3D generation and video synthesis constitutes a natural next step. In these settings, identity preservation becomes substantially more challenging due to the introduction of additional dimensions (spatial consistency across viewpoints in 3D and temporal coherence across frames in video) requiring more robust mechanisms for maintaining concept fidelity over time and space.

Second, future work could investigate region-aware controllability by refining the proposed weighting mechanisms to operate at a finer granularity within the feature space similar to~\cite{foteinopoulou2025loratorio}. Finally, extending the evaluation to a broader range of backbone architectures would provide deeper insights into the generality of the proposed approach. Such an analysis would help disentangle limitations arising from the underlying base models from those intrinsic to multi-concept customization itself.

%% file: supplementary_sections/societal_impact.tex
\section{Societal Impact}
\label{sec:societal_impact}

Our proposed methods enhance the expressive capacity of generative image models for personalized image synthesis and customized digital content creation by enabling the coherent combination of multiple user-defined concepts through community-provided LoRA modules. This capability supports a wide range of practical applications including virtual try-on systems, story-driven image generation and the realistic modeling of human–object and human–scene interactions, fostering positive societal and creative impacts.

While generative tools provide substantial opportunities for creative expression and technological advancement, they also introduce notable risks including misuse for deceptive or manipulative content and the amplification of harmful societal biases. For instance, malicious actors could leverage such capabilities to fabricate misleading interactions involving real-world individuals, potentially deceiving audiences and eroding public trust. Moreover, unresolved questions regarding authorship and attribution remain an important concern. As our approach operates exclusively at inference time and relies solely on the composition of publicly available LoRA modules without additional training, it does not introduce environmental or ethical costs associated with model training. Nevertheless, it may still inherit and propagate biases present in the underlying pretrained models or individual LoRA modules, which can be amplified in multi-concept generation scenarios. Overall, these concerns are not unique to our approach but are broadly shared across multi-concept customization methods, as well as image generative and image editing models more generally.

Consequently, mitigating the risks of misuse should remain a key research priority in generative AI. Potential mitigation strategies include the incorporation of imperceptible watermarking in generated images to discourage unauthorized use and facilitate attribution. Additionally, standardized documentation of community-provided LoRAs and post-generation automated risk assessment could further enhance responsible and transparent deployment.

%% file: supplementary_sections/qualitative_comparisons_extras.tex
\section{Additional Qualitative Comparisons}
\label{sec:qualitative_comparisons_extras}

\begin{figure*}[t]
  \centering
  \includegraphics[width=0.8\textwidth]{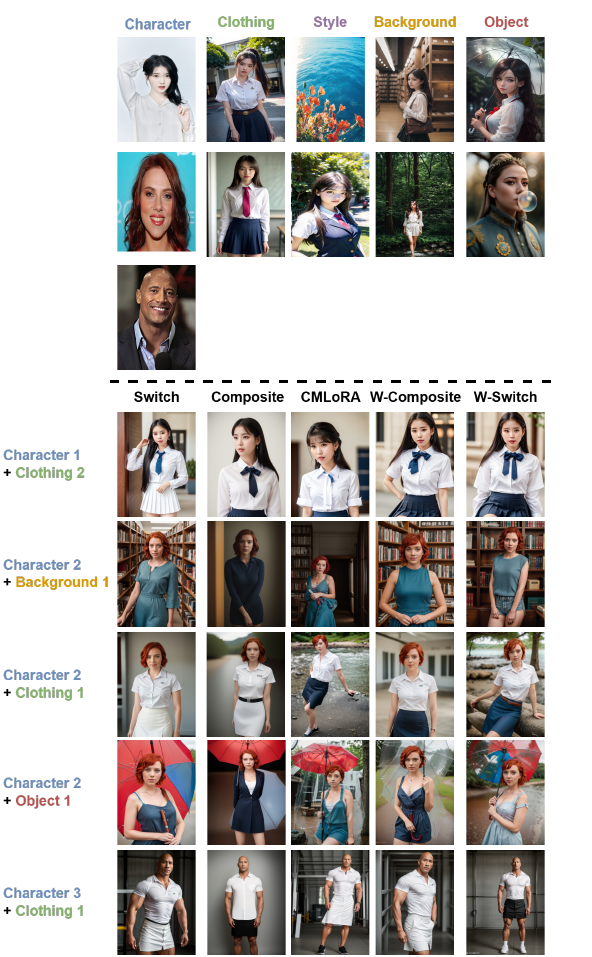}
  \caption{Examples of generated images with $N=2$ LoRA candidates across our proposed methods and baseline models in the \emph{ComposLoRA} testbed.}
  \label{fig:supplementary_grid_2}
\end{figure*}

\begin{figure*}[t]
  \centering
  \includegraphics[width=0.8\textwidth]{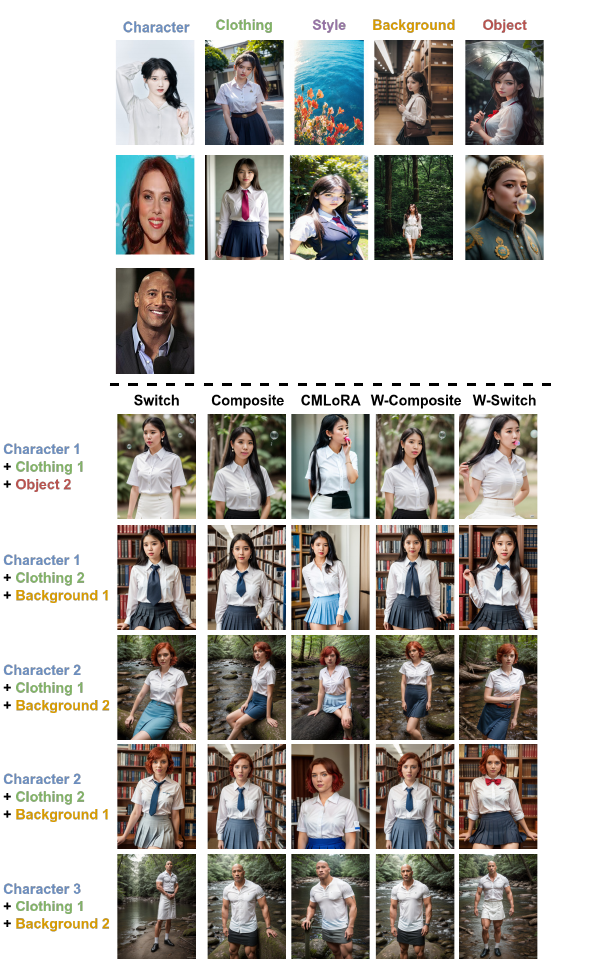}
  \caption{Examples of generated images with $N=3$ LoRA candidates across our proposed methods and baseline models in the \emph{ComposLoRA} testbed.}
  \label{fig:supplementary_grid_3}
\end{figure*}

\begin{figure*}[t]
  \centering
  \includegraphics[width=0.8\textwidth]{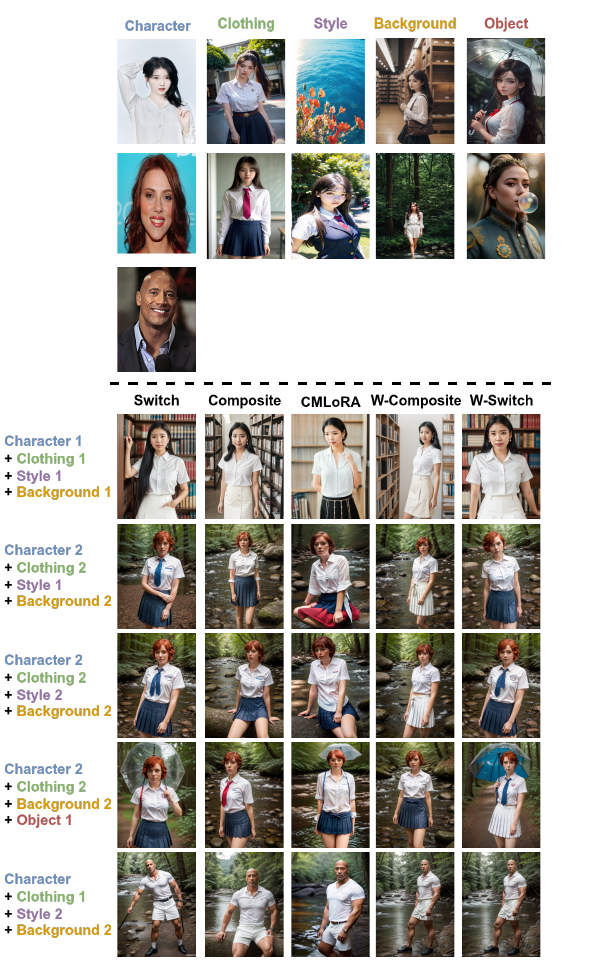}
  \caption{Examples of generated images with $N=4$ LoRA candidates across our proposed methods and baseline models in the \emph{ComposLoRA} testbed.}
  \label{fig:supplementary_grid_4}
\end{figure*}

\begin{figure*}[t]
  \centering
  \includegraphics[width=0.8\textwidth]{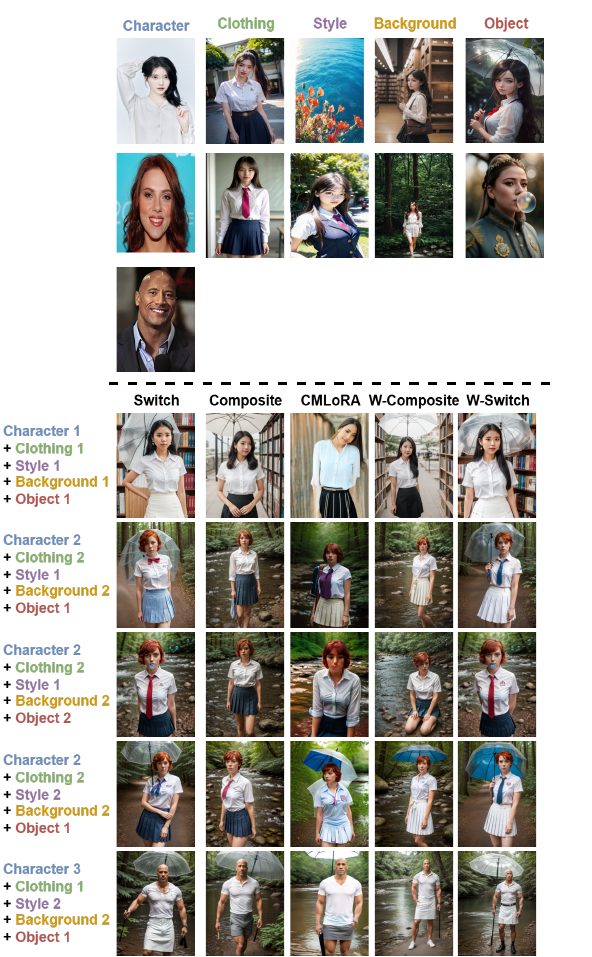}
  \caption{Examples of generated images with $N=5$ LoRA candidates across our proposed methods and baseline models in the \emph{ComposLoRA} testbed.}
  \label{fig:supplementary_grid_5}
\end{figure*}

In this section, we present additional qualitative results showcasing images generated using \emph{W-Switch} and \emph{W-Composite} and compare them against the three examined state-of-the-art baselines. Figs.~\ref{fig:supplementary_grid_2}–\ref{fig:supplementary_grid_5} illustrate generations obtained from different combinations of LoRA adapters, with the number of composed LoRAs increasing from two to five. Overall, \emph{W-Switch}, followed by \emph{W-Composite}, produces images of high visual quality with minimal concept interference and little to no concept vanishing across diverse concept combinations, with these benefits becoming increasingly pronounced as the number of composed LoRAs grows.

